\pdfoutput=1
\documentclass[10pt,twocolumn,letterpaper]{article}
\usepackage[pagenumbers]{cvpr}
\usepackage{caption}
\usepackage{amsmath,amssymb,amsthm,mathtools}
\usepackage{booktabs}
\usepackage{algorithm}
\usepackage[noend]{algpseudocode}
\usepackage{xspace}
\usepackage{placeins}
\usepackage{xurl}

\theoremstyle{plain}
\newtheorem{theorem}{Theorem}
\newtheorem{proposition}[theorem]{Proposition}

\theoremstyle{definition}

\theoremstyle{remark}

\newcommand{\AccF}{\ensuremath{\mathrm{Acc}_{50}}\xspace}
\newcommand{\Acc}{\ensuremath{\mathrm{Acc}}\xspace}

\definecolor{cvprblue}{rgb}{0.21,0.49,0.74}
\usepackage[pagebackref,breaklinks,colorlinks,allcolors=cvprblue]{hyperref}
\hypersetup{pdftitle={What Survives Label Downsampling? Exact Region Survival for Coarse Segmentation Targets},pdfauthor={Zhiyuan Zhang, Yukun Ou, San Tek Choi}}

\title{What Survives Label Downsampling?\\Exact Region Survival for Coarse Segmentation Targets}
\author{%
Zhiyuan Zhang\textsuperscript{1,$*$}\quad
Yukun Ou\textsuperscript{2,$*$}\quad
San Tek Choi\textsuperscript{3,$*$}\\[2pt]
{\normalsize \textsuperscript{1}East China Jiaotong University\quad
\textsuperscript{2}Imperial College London\quad
\textsuperscript{3}University of Macau}\\[2pt]
{\small \textsuperscript{$*$}Equal contribution. Correspondence: \texttt{2025028081100018@ecjtu.edu.cn}}%
}

\makeatletter
\g@addto@macro\@maketitle{%
  \begin{center}
  \vspace{-16pt}
  \includegraphics[width=\textwidth]{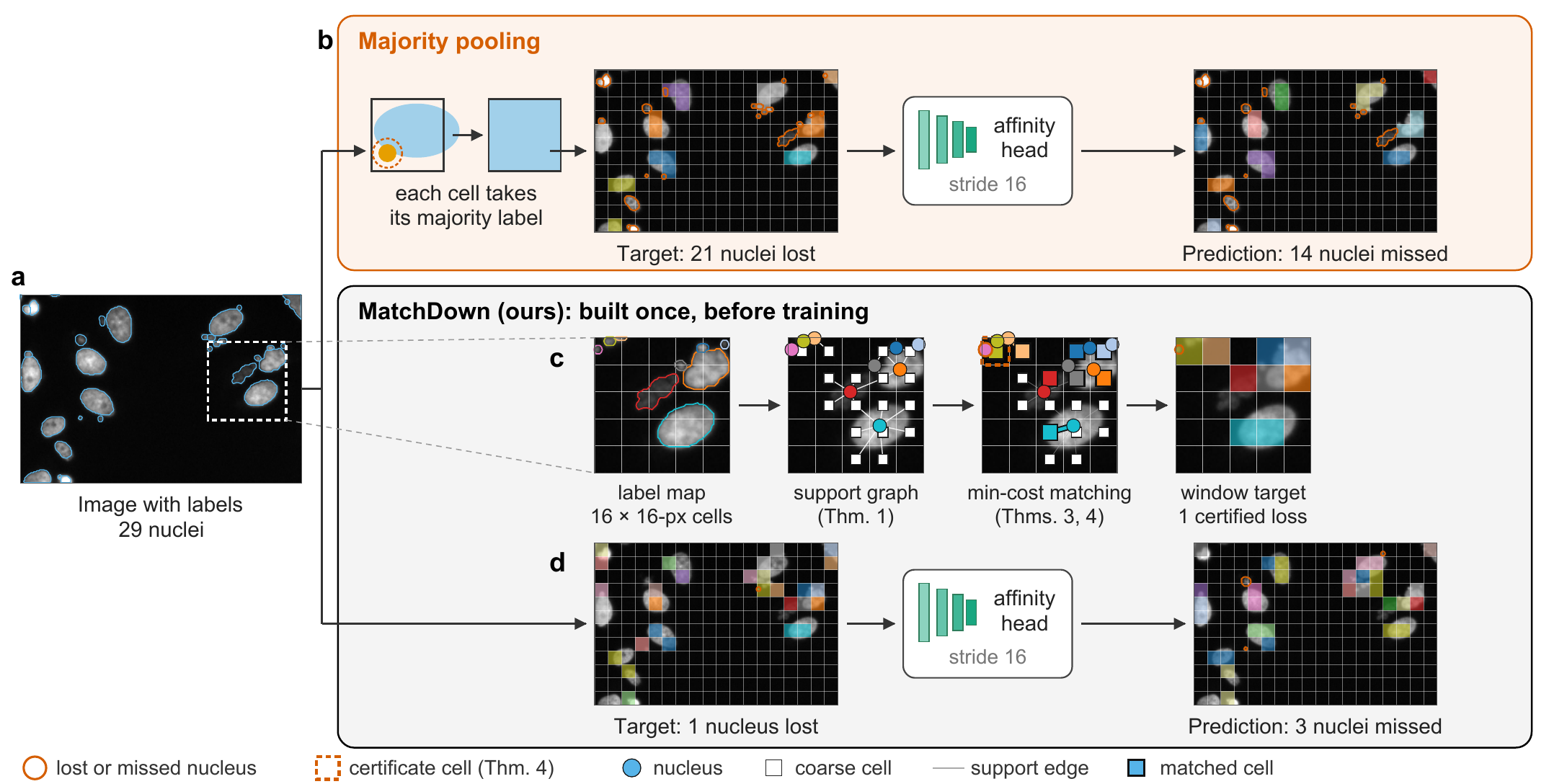}
  \captionof{figure}{\textbf{Overview.} One held-out BBBC039 crop ($192{\times}288$ pixels, 29 nuclei), downsampled by a factor of 16 and chosen because majority pooling loses many of its nuclei (10.1\% of all BBBC039 nuclei at this factor).
  (a) The image with its labels; the dashed square is the $5{\times}5$-cell window shown in (c).
  (b) Majority pooling gives each cell the label that covers most of its pixels, so a nucleus that is not the majority in any cell is missing from the target. Here 21 nuclei are missing (outlined), and the affinity head trained on this target misses 14 (seed 0, default thresholds).
  (c) MatchDown on the window: the label map on the grid; the support graph linking each nucleus to every cell it overlaps (\cref{thm:matroid}); the minimum-cost matching (filled cells, \cref{thm:assignment}), in which two nuclei share only the top-left cell, so one of them must be lost and that cell (dashed) certifies the loss (\cref{thm:certificate}); and the resulting target, which loses exactly that nucleus.
  (d) The MatchDown-P target of the whole crop loses one nucleus, and the head trained on it misses 3.}
  \label{fig:teaser}
  \vspace{6pt}
  \end{center}
}
\makeatother

\begin{document}
\raggedbottom
\maketitle
\begin{abstract}
Segmentation networks that predict at an output stride are trained against label maps downsampled by strided sampling, nearest-neighbor resampling or majority pooling.
These methods label each coarse grid cell on its own, so a region that wins no cell is absent from the target and receives no positive label.
We present MatchDown, which builds the coarse target from a minimum-cost matching between regions and the grid cells they occupy.
It keeps as many regions as any one-region-per-cell target can, certifies how many losses are unavoidable, and with priority tiers (MatchDown-P) keeps every evaluated region (a region that a metric scores) of at least one cell's area.
On COCO, Cityscapes and BBBC038 at factors 2 to 16, MatchDown-P loses exactly the minimum number of regions, whereas the downsampling used by StarDist, nnU-Net and connectomics pyramids loses four to five times as many on COCO at factor 8.
Keeping a region in the target and reconstructing its shape are different problems: we show that no region lost by majority pooling can be reconstructed above IoU 0.5 by upsampling any coarse target, yet on BBBC038 a star-convex head trained on MatchDown targets segments 35\% of the nuclei that majority pooling loses above IoU 0.5, against 15\% when trained on majority targets and 24\% on StarDist's strided target.
On two nucleus datasets, with thresholds cross-validated for count error, this head has instance accuracy at IoU 0.5 (\AccF) 2.6 to 4.2 points above the strided target and at most 2.8 points below cell-center sampling, and a lower count error than the strided and cell-center targets on BBBC039.
On BBBC039, an affinity head, which outputs one label per cell, lowers count error relative to majority targets by 9.3 nuclei per image at 7.2 points lower \AccF.
MatchDown runs once, before training, in place of the label-downsampling step.
\end{abstract}

\section{Introduction}
\label{sec:intro}

Many segmentation networks learn from labels at a coarser resolution than the image.
Grid-based object heads such as StarDist build their target by sampling the label map with a stride, \texttt{lbl[::f, ::f]} \cite{schmidt2018stardist}, and nnU-Net v2 resamples the labels of its deep-supervision heads with nearest-exact interpolation \cite{isensee2021nnunet,batchgeneratorsv2}.
DeepLab subsampled its ground truth by 8 \cite{chen2017deeplab}, and connectomics pyramids pool labels $2{\times}2$ per level with the mode \cite{tinybrain}.
Each of these methods labels every coarse grid cell independently of the others, so a region that is not selected in any cell it touches is missing from the target (\cref{fig:teaser}).
These losses are not limited to tiny regions: at factor 8, the standard methods lose 150 to 195 regions of a cell or more on 1,000 COCO panoptic images, and majority pooling loses 8.8\% of the nuclei of BBBC038 (\cref{sec:exp:label}).

For some heads the coarse target cannot be avoided.
Grid-based instance heads such as StarDist, CenterNet and SOLO assign one object to each output cell \cite{schmidt2018stardist,zhou2019objects,wang2020solo}, StarDist's 3D models predict on a subsampled grid to fit volumes in memory \cite{weigert2020star}, deep supervision needs a label at every scale \cite{isensee2021nnunet}, and label pyramids are built once and reused \cite{tinybrain}.
Patch-token heads of vision transformers \cite{dosovitskiy2021vit,oquab2024dinov2} face the same choice whenever their loss is computed on the token grid rather than after upsampling.
A region missing from the coarse target receives no positive label in that image; a network may still find it by generalizing from other images, but it is not trained to.

A coarse target faces two separate limits: whether a region is assigned a cell at all, and how well whole cells reproduce its shape.
A $2{\times}2$ nucleus inside an $8{\times}8$ cell of background is lost by majority pooling; kept, its whole-cell mask reaches an IoU of only $4/64$.
Both fail an IoU-0.5 test, but only the kept nucleus gives a shape-predicting head a cell from which to regress its outline.
Small size alone does not force a loss: tiny regions survive whenever enough distinct cells are available, whereas of $k$ regions that lie inside a single cell and touch no other, at most one can be kept.
Keeping every region is therefore not always possible, because several regions can compete for the same few cells.
Soft-label downsampling \cite{alcover2026soft} keeps the class proportions of each cell but does not specify which regions are kept.
Greedy rescue recovers about half of the lost regions on COCO and nearly all lost nuclei, but it has no guarantee and can keep as few as $k{+}1$ of $4k{+}1$ regions (\cref{sec:method:rescue}).
Chen and Peng's topology-preserving integer program \cite{chen2024topology} handles binary masks and can be infeasible.
None of these methods determines how many regions can be kept on a given grid, which losses are unavoidable, or which target attains the maximum.

We answer these questions with MatchDown (matching-based downsampling), which builds coarse targets from the support graph between regions and the grid cells that contain them (\cref{fig:teaser}c).
With priority tiers (MatchDown-P), it keeps every evaluated region of at least one cell's area and loses only the minimum number of evaluated regions.
Our analysis has three parts.
\emph{Exact survival:} the sets of regions that can be kept together are the independent sets of a transversal matroid, and every region of at least one cell's area can be kept, because $k$ such regions touch at least $k$ cells.
A single minimum-cost matching attains the maximum with the least pixel disagreement.
\emph{Lower bound and certificates:} every unavoidable loss is certified by a cluster of $k$ regions that share fewer than $k$ cells.
\emph{Costs:} a coverage threshold admits the same exact count, a per-region bound limits the IoU that any coarse target can reach (no region that majority pooling loses can exceed IoU 0.5), and a constrained repair removes some of the spurious adjacencies that keeping more regions introduces.
We build on classical results \cite{hall1935,edmonds1965transversals,dulmage1958coverings} and solve with standard minimum-cost matching \cite{jonker1987shortest}; our contribution is the formulation and its guarantees.

On COCO, Cityscapes and BBBC038 at factors 2 to 16, MatchDown-P loses exactly the minimum number of evaluated regions and none of a cell or more, while the standard methods lose four to five times the minimum on COCO at factor 8 (\cref{fig:audit}).
The kept regions are reconstructed coarsely: on COCO-R at factor 8, the share of reachable regions reconstructed at IoU 0.5 falls from 98.4\% for majority pooling to 89.5\% (\cref{tab:audit}), and none of the regions that majority pooling loses can be reconstructed above IoU 0.5 by upsampling any coarse target (\cref{prop:ceiling}).
The kept regions still help a shape-predicting head: trained on MatchDown targets, a star-convex head segments 35\% of the BBBC038 nuclei missing from the majority target above IoU 0.5, against 15\% on majority targets and 24\% on StarDist's strided target.
With cross-validated thresholds it has the lowest count error of the targets in \cref{tab:headline}, with \AccF 2.6 to 4.2 points above the strided target and at most 2.8 below cell-center sampling; on BBBC038 the strided head has the lower count error when the NMS range is extended to 0.1 (\cref{tab:nms}).
On BBBC039 at factor 16, an affinity head, which outputs one label per cell, detects 30\% of the nuclei missing from the majority target, against 3\% for the head trained on that target, and lowers count error by 9.3 nuclei per image at 7.2 points lower \AccF.
On BBBC038 at factor 8, greedy rescue and a full-resolution loss match MatchDown's count reduction, the latter at higher \AccF; building a MatchDown target takes 1.7 to 1.9 times as long as majority pooling.

Our contributions are summarized as follows:
\begin{itemize}
\item \textbf{Exact region survival on a fixed coarse grid:} a matching formulation of label downsampling with maximum-survival targets at minimum disagreement, loss certificates and the one-cell guarantee.
\item \textbf{Assignment versus reconstruction:} a per-region IoU bound that no region lost by majority pooling can exceed, and evidence that a shape-predicting head recovers such regions beyond it.
\item \textbf{A label-level evaluation and training results} against the downsampling of StarDist, nnU-Net and connectomics pyramids, with two heads on two nucleus datasets.
\end{itemize}
Code and all coarse targets will be released.

\section{Related Work}
\label{sec:related}

\paragraph{Heads that train from coarse targets.}
Fully convolutional networks predict at a stride and upsample \cite{long2015fully}; DeepLab computed its loss against ground truth subsampled by 8 \cite{chen2017deeplab}.
Dense detectors assign objects to the cells of a strided grid: YOLO \cite{redmon2016yolo} and CenterNet \cite{zhou2019objects} give each object one cell, FCOS \cite{tian2019fcos} assigns the cells inside a box with a rule for cells claimed by several boxes, and SOLO \cite{wang2020solo} predicts one instance per grid cell; CenterNet reports how often two object centers fall into the same output cell on COCO.
ATSS \cite{zhang2020atss}, OTA \cite{ge2021ota} and the bipartite matching of DETR \cite{carion2020detr} choose during training which predictions are supervised by which objects.
MatchDown acts on the label map instead: before training, it decides which regions a one-region-per-cell target can contain.
StarDist \cite{schmidt2018stardist,weigert2020star} trains from per-cell targets, and SSAP predicts affinities at strides 4 to 64 \cite{gao2019ssap}.
Affinity- and watershed-based decoders \cite{turaga2009maximin,funke2019large,wolf2018mutex,bai2017deep}, U-Net \cite{ronneberger2015unet}, Cellpose \cite{stringer2021cellpose}, HoVer-Net \cite{graham2019hover} and Omnipose \cite{cutler2022omnipose} are usually trained at full resolution, and Mask2Former computes its mask loss at points sampled from the full-resolution label \cite{cheng2022mask2former}; none of these needs a coarse target.

\paragraph{Soft and full-resolution targets.}
Soft-label downsampling replaces the hard coarse label by the class proportions of each cell \cite{alcover2026soft}.
Soft labels keep class information that a hard argmax discards, but they do not distinguish two regions of the same class or specify which regions remain represented.
We produce hard, region-valued targets with a guarantee on which regions are kept, and \cref{sec:exp:ablation} trains both kinds on the same images.

\paragraph{Topology-preserving downsampling and representations.}
Chen and Peng \cite{chen2024topology} downsample binary images with an integer program.
Whenever the program is feasible, it keeps every component and the region adjacency graph, and hence $b_0$ and $b_1$.
Its constraints give every component a cell of its own, so it is infeasible whenever Hall's condition fails on its support graph; our reimplementation returns no solution in 22 of 300 cases (21 infeasible, one time-out), four of them for this reason (\cref{sec:supp:labelcmp}).
We address partitions with many labels and always return a labeling, but we guarantee less: maximum survival, and a repair that never increases the number of coarse components of a region; we guarantee neither holes nor exact adjacency (\cref{sec:method:scope}).
SEMIR \cite{miller2026semir} and topology-preserving transformations \cite{ngo2014topology,passat2022homotopic,decenciere2007adaptive} change the lattice or preserve topology under resampling, without an objective on which regions are kept; we keep the regular coarse grid that standard networks predict on.

\paragraph{Assignment to grid cells.}
Matching regions to cells appears in grid-map layout \cite{eppstein2015improved}, where regions may be placed in any cell rather than only in cells they occupy.
Our survival theorems apply Hall's theorem, the transversal-matroid theorem and the Dulmage--Mendelsohn decomposition \cite{hall1935,edmonds1965transversals,dulmage1958coverings,konig1931graphok,ore1955graphs,hopcroft1973n,lovasz1986matching,oxley2011matroid} to the support graph of a downsampling, and we solve the resulting problems with standard minimum-cost matching algorithms \cite{kuhn1955hungarian,munkres1957algorithms,jonker1987shortest,crouse2016implementing}.

\section{Exact Region Survival}
\label{sec:method}
MatchDown takes three inputs: a label map, the factor $f$ at which the loss is computed, and the labels that a metric scores.
It returns the coarse target that the loss is computed against, together with a certificate for the regions that no coarse target could have kept (\cref{fig:teaser}).
It runs once, before training, in place of the label-downsampling step, and the network, its loss at stride $f$ and inference do not change.
\Cref{alg:matchdown} in the supplement lists the construction.

\subsection{Problem}
\label{sec:method:problem}
A source label map on a grid of $P$ pixels is split into $R$ \emph{regions}, the 4-connected components of its labels; two pieces of one label are two regions.
A downsampling factor $f$ partitions the grid into $C$ coarse \emph{cells} of $f{\times}f$ pixels, clipped at the border.
Let $n_{rc}\ge 0$ be the number of pixels of region $r$ in cell $c$, and $|c|$ the cell's pixel count.
The \emph{support graph} $G$ is bipartite between regions and cells, with an edge $(r,c)$ whenever $n_{rc}>0$ (\cref{fig:teaser}c).
A coarse labeling $a$ gives each cell one region of its support, $n_{a(c),c}>0$, and region $r$ \emph{survives} if some cell carries it.
We measure survival $K(a)=|\{a(c)\}|$ and pixel disagreement $D(a)=P-\sum_c n_{a(c),c}$, the source pixels whose region differs from their cell's after nearest upsampling.
Spurious and missing adjacencies $F(a)$ and $L(a)$ count the region pairs that are 4-adjacent on the coarse grid but not in the source, and the reverse; $b_r(a)$ is the number of coarse components of region $r$.
Experiments count \emph{evaluated} regions, those of the labels a metric scores (not void or crowd); the theorems count all regions.

\subsection{Survivable sets and the one-cell guarantee}
\label{sec:method:matroid}
\begin{theorem}[Matroid structure]
\label{thm:matroid}
A set $S$ of regions can survive together in some labeling if and only if $G$ has a matching that saturates $S$.
The survivable sets are the independent sets of the transversal matroid of $G$ on the regions.
Hence every maximal survivable set has $\nu(G)$ regions, $\nu(G)$ the maximum matching size, and for any ranking of the regions into priority tiers the greedy rule (add the next region, in tier order, if the set stays survivable) returns a survivable set that, for every $t$, contains the maximum possible number of regions of the $t$ highest priority tiers.
\end{theorem}
\noindent
The first sentence holds because surviving regions need distinct cells, and any matching extends to a labeling by filling the unmatched cells with any region of their support.
The rest is the transversal-matroid theorem of Edmonds and Fulkerson \cite{edmonds1965transversals} and the greedy property of matroids.
Our construction, \emph{MatchDown-P} (P for priority), ranks regions in four tiers: evaluated regions of at least $f^2$ pixels ($\tau{=}4$), smaller evaluated regions ($\tau{=}3$), non-evaluated regions of at least $f^2$ pixels ($\tau{=}2$) and smaller non-evaluated regions ($\tau{=}1$); $f^2$ pixels is the area of one cell, from which \cref{thm:onecell} guarantees survival.
It solves the minimum-cost matching of \cref{thm:assignment} with every real edge of region $r$ raised by $B(4-\tau(r))$ and the dummy cost set to $4B{+}1$ (\cref{alg:matchdown}), so keeping a region of tier $\tau$ saves about $B\tau$.
Because the survivable sets form a matroid, the result keeps the largest possible number of tier-4 regions, of tier-4 and tier-3 regions together, and so on (\cref{thm:matroid}, \cref{prop:tiers}); because $B$ exceeds any total pixel cost, disagreement is minimized only among these sets.

\begin{theorem}[One-cell survival]
\label{thm:onecell}
Some labeling with maximum survival $K(a)=\nu(G)$ keeps every region of at least $f^2$ pixels.
\end{theorem}
\noindent
Hall's condition holds for the large regions because $k$ of them hold at least $kf^2$ pixels and a cell holds at most $f^2$, so they touch at least $k$ cells \cite{hall1935}; the matching that saturates them extends to one of size $\nu(G)$ by matroid augmentation (\cref{thm:matroid}, \cref{sec:supp:proofs}).
None of the per-cell methods we compare (majority, nearest, strided, interpolation followed by argmax) has this property.
They lose regions of any size when a larger region dominates every cell the smaller one touches, and \cref{sec:exp:label} counts how many.
MatchDown-P puts the large evaluated regions in the top tier, so it keeps all of them.
The unweighted construction, MatchDown, has no such guarantee; it is the target we train on BBBC038, where both forms lose no nucleus at factor 8 (\cref{tab:bbbc038star}).

\subsection{Exact survival by matching, and its certificate}
\label{sec:method:assignment}
\begin{theorem}[Exact survival]
\label{thm:assignment}
Every labeling has $K(a)\le\nu(G)$.
Give each region a private dummy cell, set $B=P+1$ and $\ell(r,c)=\max_{r'}n_{r'c}-n_{rc}$, weight real edges $1+\ell(r,c)$ and dummy edges $B+1$, solve a minimum-weight full matching of the regions \cite{jonker1987shortest}, fill every cell with its majority region and overwrite each matched cell with its region.
The result $a_*$ attains $K(a_*)=\nu(G)$ and minimizes $D(a)$ among all labelings with $K(a)=\nu(G)$.
\end{theorem}
\noindent
The dummy penalty exceeds any total cost $\sum\ell$, so a minimum-weight full matching has exactly $\nu(G)$ real edges and, among those, the least cost.
Majority completion adds no cost, and it cannot reintroduce an unmatched region without exceeding $\nu(G)$; the full proof is in \cref{sec:supp:proofs}.
With tier weights the same construction gives MatchDown-P; with equal weights it gives the unweighted MatchDown.

\begin{theorem}[Lower bound and certificates]
\label{thm:certificate}
The number of regions any labeling loses is at least the deficiency $\delta(G)=\max_{S}\big(|S|-|N(S)|\big)$ over region sets $S$ with neighborhood $N(S)$, and the matching of \cref{thm:assignment} loses exactly $\delta(G)$.
The regions reachable by alternating paths from the unmatched regions of a maximum matching, with their cells, split into connected clusters, each of $k_i$ regions in $m_i<k_i$ cells, and $\sum_i(k_i-m_i)=\delta(G)$.
\end{theorem}
\noindent
The first sentence is the K\H{o}nig--Ore formula \cite{konig1931graphok,ore1955graphs}, and the second is the Dulmage--Mendelsohn decomposition \cite{dulmage1958coverings}.
Each cluster is a certificate that can be checked by counting: its $k_i$ regions share $m_i$ cells, so they cannot all survive and at least $k_i-m_i$ of them are lost, although the certificate does not single out which (\cref{fig:teaser}c; a worked $12{\times}12$ example is in \cref{fig:matching}).
The same statement holds for the evaluated regions on the subgraph they induce, and \cref{sec:exp:label} shows that MatchDown-P's losses equal this lower bound on every dataset and factor.

\subsection{Adjacency repair and forced adjacencies}
\label{sec:method:repair}
Starting from $a_*$, a repair sweeps the cells at most twice.
It relabels a cell only to a region of its support that already survives elsewhere, only if the old region keeps another cell and neither region gains a 4-connected coarse component, only if the disagreement stays within $D(a_*)+\lfloor P/200\rfloor$ (half a percent of the pixels), and only if $(F,L)$ decreases lexicographically (\cref{prop:repair}).
The set of surviving regions is therefore unchanged, no $b_r$ increases, and at most $2C$ cells are relabeled.
Maximum survival can force spurious adjacencies: on the $4{\times}4$ source of \cref{fig:forced}, the only labeling that keeps all four regions has two (\cref{prop:forced}).

\subsection{Survival under a coverage threshold}
\label{sec:method:kappa}
Keeping a region as small as one pixel per cell inflates it up to $f^2$-fold.
A coverage threshold $\kappa$ assigns a region only to cells it covers by at least a fraction $\kappa$, and the same exact count holds on the thresholded support graph.
The number of regions kept through such cells is at most the matching number of the thresholded graph, the construction of \cref{thm:assignment} attains it with the least disagreement, and a tiered form that ranks the regions already kept by uncovered cells below all others attains the minimum loss of evaluated regions (\cref{thm:kappa}, \cref{sec:supp:kappatheory,sec:supp:kappa}).

\subsection{A per-region IoU upper bound}
\label{sec:method:ceiling}
Instance metrics count a predicted segment as correct only if its IoU with a ground-truth segment exceeds 0.5, and a coarse target reconstructed by nearest upsampling gives each kept region whole cells.
\begin{proposition}[Per-region IoU bound]
\label[proposition]{prop:ceiling}
Under any one-region-per-cell target reconstructed by nearest upsampling, region $O$ reaches at most
\[
\mathrm{IoU}^*(O)=\max_{S\subseteq\mathrm{cells}(O)}\ \frac{\sum_{c\in S} n_{Oc}}{|O|+\sum_{c\in S}\big(|c|-n_{Oc}\big)},
\]
and the maximum is attained by a prefix of $O$'s cells sorted by $n_{Oc}/(|c|-n_{Oc})$ in decreasing order.
If $|O|\le f^2/2$ and $\hat O$ contains at least one full $f{\times}f$ cell, $\mathrm{IoU}(O,\hat O)\le 1/2$.
\end{proposition}
\noindent
With $\lambda^*$ the optimum, a cell raises the ratio exactly when $n_{Oc}>\lambda^*(|c|-n_{Oc})$, so the optimum takes every cell above $\lambda^*$ and none below; the sub-cell case follows from $|O\cap\hat O|\le f^2/2$ and $|O\cup\hat O|\ge f^2$.
The bound depends only on the grid and the region, so it holds for every coarse target; the IoU of a network that reproduces its target is bounded in the same way (\cref{fig:audit}b).
We call a region \emph{reachable} if $\mathrm{IoU}^*(O)>0.5$.
Majority pooling keeps every reachable region: $\mathrm{IoU}^*(O)>0.5$ and $|O|\ge\sum_{c\in S}n_{Oc}$ give $\sum_{c\in S}n_{Oc}>\sum_{c\in S}(|c|-n_{Oc})$, so some cell has $n_{Oc}>|c|/2$ and is won by $O$.
Hence no region that majority pooling loses is reachable, on any image and at any factor.
The bound concerns targets reconstructed cell by cell; a decoder that predicts sub-cell geometry, such as a star-convex polygon, is not limited by it.

\subsection{Greedy rescue, complexity and scope}
\label{sec:method:rescue}
A simpler alternative, \emph{greedy rescue}, starts from majority pooling and lets each lost region, smallest first, take its best-covered cell unless this removes the region currently in that cell.
Any rescue that continues until no lost region has a support cell that a surviving region can spare keeps at least $\nu(G)/2$ regions; \emph{maximal rescue}, which gives each lost region its best-covered spare cell, is such a rescue, while greedy rescue is not and can keep only $k{+}1$ of $4k{+}1$ regions (\cref{prop:rescue}, \cref{sec:supp:gadget}).
\label{sec:method:scope}
The assignment is one rectangular minimum-weight full matching, and the certificate one maximum matching with a breadth-first search; a full $1024{\times}2048$ Cityscapes frame at factor 8 takes 1.29\,s on one CPU core (\cref{tab:runtime}).
The method always returns a labeling, but it guarantees neither holes, homotopy type, the source adjacency graph nor 8-connected counts; \cref{sec:supp:checks} describes how we verified the implementation.

\section{Experiments}
\label{sec:experiments}

\subsection{Setup}
\label{sec:exp:protocol}
\paragraph{Data and regions.}
We use BSDS500 \cite{arbelaez2011contour}, three disjoint sets of 1,000 COCO panoptic validation images \cite{kirillov2019panoptic,lin2014coco} (COCO-A, COCO-B and a held-out set, COCO-R), and Cityscapes with its fine annotations \cite{cordts2016cityscapes}.
The nucleus datasets are BBBC038 \cite{caicedo2019nucleus} and BBBC039 \cite{ljosa2012annotated,caicedo2019evaluation}.
Regions are 4-connected components of instances (things) and of classes (stuff); the evaluated regions are those of the labels that a metric scores, \ie, not void or crowd.
\Cref{tab:datasets} in the supplement lists the dataset statistics.

\paragraph{Baselines.}
We compare MatchDown with the downsampling methods of existing pipelines: strided sampling (StarDist's target, and PyTorch's default nearest interpolation), nearest-exact (nnU-Net v2's deep supervision), nearest sampling at the cell center, majority pooling (the mode, as in connectomics pyramids), max pooling, one-hot bilinear interpolation with argmax, and nnU-Net's per-label interpolation \cite{isensee2021nnunet}.
We also compare soft labels \cite{alcover2026soft}, greedy and maximal rescue, a full-resolution loss where one can be computed, and, at the label level, Chen and Peng's integer program \cite{chen2024topology}.
MatchDown denotes the unweighted construction (\cref{thm:assignment} with the repair) and MatchDown-P its tiered form (\cref{alg:matchdown}).

\paragraph{Training.}
We train two heads on the 670 BBBC038 images at factor 8, and on the 197 BBBC039 images at factor 16 (\cref{tab:headline}); the factor is the output stride of the head, which StarDist calls the grid.
Each head is trained with nine seeds and three folds, so every image is predicted by a model that never saw it, and we score at full resolution. Some BBBC038 models come from a second set of runs with the same protocol (App.~\ref{sec:supp:training}); every difference is computed within one set.
The \emph{affinity} head (ResNet-18 FPN \cite{he2016deep,lin2017feature}) predicts foreground and same-nucleus affinities at the target stride and is decoded by thresholding and connected components.
The \emph{star-convex} head predicts an object probability and 32 radial distances in each grid cell, as StarDist does \cite{schmidt2018stardist} (our reimplementation), and is decoded by non-maximum suppression.
Both heads require a coarse target, because each grid cell carries one object. We name each model by its head and target (\eg, the MatchDown affinity head); \cref{tab:hyper} lists the hyper-parameters.

\paragraph{Metrics.}
We report count error (the absolute difference between the predicted and true numbers of nuclei per image), the centroid-based F1 of the predicted objects, F1 at IoU${}>0.5$, and instance accuracy $\mathrm{Acc}_\tau=\mathrm{TP}/(\mathrm{TP}+\mathrm{FP}+\mathrm{FN})$ with one-to-one matching at IoU${}>\tau$, computed per image at a fixed decoder operating point and averaged over images. This is the accuracy measure of Weigert et al.~\cite{weigert2020star}, which Schmidt et al.~\cite{schmidt2018stardist} report as AP.
\AccF is $\mathrm{Acc}_{0.5}$, \Acc the mean over $\tau=0.50,0.55,\dots,0.95$, and differences in \AccF are given in points (0.01).
We also report the share of the nuclei missing from the majority target that a model detects or matches (\cref{sec:supp:metrics}), at the default thresholds and at thresholds tuned by three-fold cross-validation to minimize count error (\emph{cross-validated} thresholds, \cref{sec:supp:training}).
Differences come with 95\% confidence intervals from a bootstrap over seeds and images \cite{efron1979bootstrap,diciccio1996bootstrap}.
We chose BBBC039 over NuInsSeg \cite{mahbod2024nuinsseg} as a stress test, by a rule fixed before training: the largest difference in oracle count error between the majority and MatchDown-P targets (\cref{sec:supp:data}).

\subsection{Comparison with standard downsampling}
\label{sec:exp:label}
\begin{table}[t]
\centering\footnotesize
\caption{\textbf{Label-level evaluation on COCO-R at factor 8} (1,000 held-out images, 25,963 evaluated regions). Lost: evaluated regions missing from the target; $\ge$1 cell: lost regions of at least $f^2$ pixels; Recon.: share (\%) of the reachable regions (\cref{prop:ceiling}) reconstructed at IoU${}>0.5$ by nearest upsampling. The lower bound (\cref{thm:certificate}) is 1,854. Keeping the small regions costs reconstruction quality (\cref{tab:cocoa_pq}). All factors and datasets: \cref{tab:supp:lost,tab:supp:f8}. Best in bold, second best underlined.}
\label{tab:audit}
\setlength{\tabcolsep}{4pt}
\begin{tabular}{@{}lrrc@{}}
\toprule
Target & Lost $\downarrow$ & $\ge$1 cell $\downarrow$ & Recon. $\uparrow$ \\
\midrule
Majority pooling & 9,207 & 195 & \textbf{98.4} \\
Strided (StarDist) & 8,498 & 176 & 81.1 \\
Nearest-exact (nnU-Net) & 8,481 & 150 & \underline{97.7} \\
Max pooling & 7,849 & 933 & 74.5 \\
Greedy rescue & 4,921 & 69 & 89.9 \\
Maximal rescue & 4,381 & \underline{9} & 89.6 \\
MatchDown (ours) & \underline{3,185} & \textbf{0} & 89.5 \\
MatchDown-P (ours) & \textbf{1,854} & \textbf{0} & 89.5 \\
\bottomrule
\end{tabular}
\end{table}

\begin{figure}[t]
\centering
\includegraphics[width=\columnwidth]{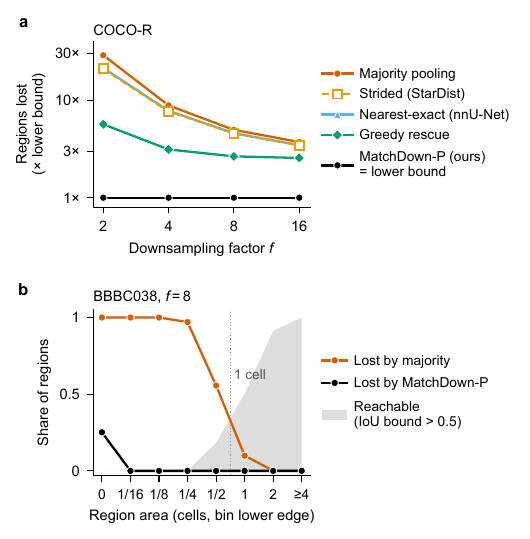}
\caption{\textbf{Regions lost against the lower bound, and loss by nucleus size.} (a) Evaluated regions lost on COCO-R as a multiple of the lower bound of \cref{thm:certificate}. At factor 2, majority pooling loses 29 times the minimum and strided and nearest-exact sampling (which nearly coincide) 21 times; at factors 8 and 16 they lose 3.5 to 5 times, greedy rescue 2.6 to 5.7 times across factors, and MatchDown-P exactly the minimum. (b) BBBC038 at $f{=}8$, by region area (nuclei and their 4-connected pieces; bins labeled by their lower edge; dotted line: one cell): the share of regions lost by majority pooling and by MatchDown-P (38 pieces below a sixteenth of a cell, all certified; MatchDown-P loses no nucleus), and the share that is reachable, \ie, whose IoU bound exceeds 0.5 (\cref{prop:ceiling}, shaded). Other datasets: \cref{fig:auditsupp,fig:factorsupp}.}
\label{fig:audit}
\end{figure}

\Cref{thm:certificate} lets us measure how far each method is from the minimum loss. \Cref{tab:audit} counts the evaluated regions each method loses on 1,000 held-out COCO images at factor 8, and \cref{fig:audit}a extends the comparison to every factor.
At factor 8, majority pooling, strided and nearest-exact sampling lose 4.6 to 5.0 times the minimum.
Each of them loses regions of a cell or more, and greedy rescue halves the loss.
MatchDown-P loses exactly the lower bound at every factor, as \cref{prop:tiers} requires, and never loses a region of a cell or more.
The same holds for all 20 dataset--factor pairs (\cref{tab:supp:lost,tab:supp:f8}), and the three standard methods lose regions of a cell or more on each.

\subsection{Effect on trained networks}
\label{sec:exp:training}
\begin{table*}[t]
\centering\footnotesize
\caption{\textbf{Main results} with cross-validated thresholds: count error (CE, nuclei per image), F1 at IoU${}>0.5$ and \AccF of both heads on both datasets; BBBC038 (670 images) at factor 8, BBBC039 (197 images) at factor 16 and, for the star-convex head, also at factor 8; 9 seeds $\times$ 3 folds. $^\ast$MatchDown on BBBC038 and MatchDown-P on BBBC039 (on BBBC038 both lose no nucleus). Best of the targets shown in bold (ties share the mark), second underlined. Other targets: \cref{tab:ablation}; details: \cref{tab:training,tab:bbbc038star,tab:bbbc039,tab:bbbc039diff}.}
\label{tab:headline}
\setlength{\tabcolsep}{2.6pt}
\begin{tabular}{@{}lrrrrrrrrrrrrrrr@{}}
\toprule
 & \multicolumn{6}{c}{BBBC038} & \multicolumn{9}{c}{BBBC039} \\
\cmidrule(lr){2-7}\cmidrule(lr){8-16}
 & \multicolumn{3}{c}{Affinity, $f{=}8$} & \multicolumn{3}{c}{Star-convex, $f{=}8$} & \multicolumn{3}{c}{Affinity, $f{=}16$} & \multicolumn{3}{c}{Star-convex, $f{=}16$} & \multicolumn{3}{c}{Star-convex, $f{=}8$} \\
\cmidrule(lr){2-4}\cmidrule(lr){5-7}\cmidrule(lr){8-10}\cmidrule(lr){11-13}\cmidrule(lr){14-16}
Target & CE~$\downarrow$ & F1~$\uparrow$ & \AccF~$\uparrow$ & CE~$\downarrow$ & F1~$\uparrow$ & \AccF~$\uparrow$ & CE~$\downarrow$ & F1~$\uparrow$ & \AccF~$\uparrow$ & CE~$\downarrow$ & F1~$\uparrow$ & \AccF~$\uparrow$ & CE~$\downarrow$ & F1~$\uparrow$ & \AccF~$\uparrow$ \\
\midrule
Majority pooling & 5.01 & \underline{0.689} & \underline{0.541} & 6.50 & 0.830 & \underline{0.750} & 13.05 & \underline{0.660} & \underline{0.497} & 13.30 & 0.897 & \textbf{0.872} & 6.80 & 0.922 & \underline{0.871} \\
Cell center & \underline{3.91} & \textbf{0.696} & \textbf{0.558} & 5.78 & \underline{0.841} & \textbf{0.770} & 10.43 & \textbf{0.698} & \textbf{0.539} & 11.57 & \textbf{0.922} & \underline{0.855} & 5.51 & \underline{0.926} & \underline{0.871} \\
Strided (StarDist) & 4.10 & 0.440 & 0.270 & \underline{4.84} & 0.809 & 0.728 & \underline{8.70} & 0.080 & 0.070 & \underline{5.62} & 0.888 & 0.799 & \underline{4.29} & 0.915 & 0.847 \\
MatchDown$^\ast$ (ours) & \textbf{3.69} & 0.673 & 0.530 & \textbf{4.10} & \textbf{0.854} & \textbf{0.770} & \textbf{3.77} & 0.591 & 0.425 & \textbf{3.35} & \underline{0.904} & 0.827 & \textbf{2.90} & \textbf{0.932} & \textbf{0.873} \\
\bottomrule
\end{tabular}
\end{table*}

\begin{table}[t]
\centering\footnotesize
\caption{\textbf{Ablation on the affinity head}, BBBC038 at factor 8; soft labels as in \cite{alcover2026soft}. Count error at the default thresholds, then, with cross-validated thresholds, the differences from the majority model and F1 at IoU${}>0.5$. $^\dagger$Masked: majority targets with the cells of the missing nuclei left out of the loss, from the second set of runs, differences against its majority model (in that set, MatchDown's difference is $-1.34$). Intervals and further columns: \cref{tab:training,tab:controls}; the star-convex head: \cref{tab:bbbc038star}. Best in bold, second best underlined.}
\label{tab:ablation}
\newlength{\ablationnumberwidth}
\newcommand{\ablationboldnum}[1]{%
  \settowidth{\ablationnumberwidth}{$#1$}%
  \resizebox{\ablationnumberwidth}{\height}{{\boldmath$#1$}}%
}
\setlength{\tabcolsep}{3pt}
\begin{tabular}{@{}lrrrr@{}}
\toprule
 & Count err.~$\downarrow$ & $\Delta$count err.~$\downarrow$ & $\Delta$\AccF~$\uparrow$ & F1~$\uparrow$ \\
Target & default & \multicolumn{3}{c}{cross-validated} \\
\midrule
Majority pooling & 5.93 & ref. & ref. & \underline{0.689} \\
Masked$^\dagger$ & 5.02 & $-0.77$ & \underline{$+0.000$} & \underline{0.689} \\
Greedy rescue & \underline{3.93} & \underline{$-1.43$} & $-0.010$ & 0.679 \\
Soft labels & 6.46 & \ablationboldnum{-1.64} & $-0.082$ & 0.598 \\
Full-res.\ loss & \textbf{3.56} & $-1.34$ & \ablationboldnum{+0.018} & \textbf{0.690} \\
MatchDown (ours) & 4.03 & $-1.32$ & $-0.011$ & 0.673 \\
\bottomrule
\end{tabular}
\end{table}

\paragraph{Star-convex head.}
\Cref{tab:headline} reports both heads on both datasets with cross-validated thresholds; \cref{tab:training,tab:bbbc038star,tab:bbbc039,tab:bbbc039diff} add the default thresholds, the differences with their intervals and the shares of missing nuclei that are detected.
On BBBC038 at factor 8 and the default thresholds, the strided head's \AccF is about ten points below those of the heads trained on cell-center, rescue and MatchDown targets (\cref{tab:bbbc038star}); cross-validated thresholds narrow the difference to 4.2 points.
Of the four BBBC038 star-convex heads in \cref{tab:headline}, the MatchDown head has the lowest count error and the highest F1 at IoU${}>0.5$, and it matches 35\% of the nuclei missing from the majority target at IoU${}>0.5$, against 24\% for the strided head.
On BBBC039 at factor 16, its count error is 4.0 nuclei per image below the strided head's at the default thresholds and 2.3 below with cross-validated thresholds, with \AccF 2.7 points higher (paired difference, \cref{tab:bbbc039diff}); majority leads in \AccF, whereas cell center leads in F1.
At factor 8 on the same images the count-error differences are smaller and every difference from the strided head has the same sign (\cref{tab:bbbc039diff}).
With cross-validated thresholds at factor 4, MatchDown also improves both count error (3.80 against 4.20) and \AccF (0.795 against 0.781) over the strided target (\cref{tab:bbbc038star}). When the NMS search range is extended to 0.1, the strided head has the lower count error at factors 8 and 4 (4.02 against 5.13, and 3.61 against 4.49), and the \AccF difference narrows to 1.9 and 0.4 points (\cref{tab:nms}).

\begin{figure*}[t]
\centering
\includegraphics[width=\textwidth]{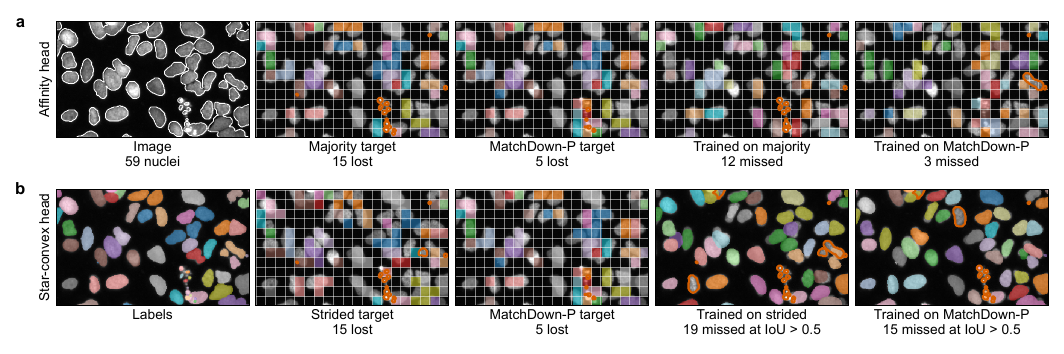}
\caption{\textbf{Qualitative results on a held-out BBBC039 image at factor 16} ($192{\times}320$-pixel window, 59 nuclei; seed 0, default thresholds). (a) Affinity head: the image, two targets with their lost nuclei outlined, and the predictions of the models trained on them, with the nuclei that no predicted object covers by a fifth outlined. (b) Star-convex head: the same for the strided and MatchDown-P targets, with the nuclei that no prediction matches at IoU${}>0.5$ outlined. More images: \cref{fig:qualsupp}.}
\label{fig:qual}
\end{figure*}

\paragraph{Affinity head.}
The affinity head is trained to predict foreground only where its target has it, so a nucleus missing from the target receives no positive label.
At the default thresholds, the affinity head trained on majority targets at factor 8 has a count error of 5.93 nuclei per image, and the one trained on MatchDown targets 4.03 (\cref{tab:training}).
Most of the difference comes from the missing nuclei (\cref{sec:exp:ceiling}): the MatchDown head detects 53\% of the nuclei missing from the majority target, against 31\% for the majority head (cross-validated thresholds, \cref{tab:training}; detection is defined in \cref{sec:supp:metrics}).
At factor 16 on BBBC039 and the default thresholds, the count error is 15.9 nuclei per image for the majority head and 3.95 for the MatchDown-P head.
This difference, 11.9 nuclei, matches the gap of 11.7 nuclei between the targets themselves, and the MatchDown-P head detects 30\% of the nuclei missing from the majority target, against 3\% (\cref{tab:bbbc039}, \cref{fig:qual}).
\AccF is lower for the MatchDown-P head, by 5.4 points at the default thresholds and by 7.2 with cross-validated thresholds (F1 0.591 against 0.660), where its count error is still 9.3 nuclei lower than the majority head's (\cref{sec:exp:ceiling}).

\subsection{Where the count reduction comes from}
\label{sec:exp:mechanism}
On BBBC038 at factor 8 and the default thresholds, the per-image count-error reduction of the MatchDown affinity head grows with the number of nuclei missing from the majority target, by 0.45 per missing nucleus (with cross-validated thresholds the interval of this slope includes zero), and the trained network recovers about half of the difference between the targets' own count errors; on BBBC039 at factor 16 it recovers all of it (\cref{fig:mechanism}).
Over the whole threshold grid, the MatchDown model has lower count error at every foreground threshold and lower \AccF at every threshold.
At thresholds that give both models the same false positives per image, the count difference is $-0.03$ $[-1.08,+0.81]$ (\cref{tab:training}), so on BBBC038 the count reduction comes with more predicted objects; selecting thresholds within each held-out fold gives the same count reduction with more false positives (\cref{sec:supp:training}).

\subsection{Assignment versus reconstruction}
\label{sec:exp:ceiling}
The nuclei that majority pooling and the other methods lose are mostly smaller than a grid cell (\cref{fig:audit}b), and by \cref{prop:ceiling} none of the regions that majority pooling loses is reachable; on BBBC038 at factor 8 the mean IoU bound of these nuclei is 0.30.
The reachable share falls below half on all three datasets at factor 16 (\cref{fig:factorsupp}b).
An affinity head outputs one label per cell, so it can detect these nuclei but not segment them at IoU${}>0.5$: 89\% of its count reduction on BBBC038 comes from nuclei that the majority model misses and the MatchDown model detects, and no affinity model matches any of them at IoU${}>0.5$.
Each such detection counts as a false positive at IoU${}>0.5$ and lowers the \AccF of the MatchDown affinity models (\cref{fig:mechanism}c).
The star-convex head decodes polygons at full resolution and is not limited by the bound: it matches more of these nuclei at IoU${}>0.5$ whenever its target keeps them, 33\% on rescue, 35\% on MatchDown and 40\% on dilated targets, against 15\% on majority and 24\% on strided targets (\cref{tab:bbbc038star}).

\subsection{Ablation study}
\label{sec:exp:ablation}
\paragraph{Components of MatchDown.}
The tiers decide which regions are kept when not all can be: on COCO-R at factor 8, MatchDown-P loses 1,854 evaluated regions, the lower bound, and the unweighted MatchDown 3,185 (\cref{tab:audit}), at a cost of at most 0.02\% of pixels (\cref{sec:supp:audit}).

\paragraph{Simpler targets.}
On BBBC038 at factor 8, greedy rescue loses 103 of the 29,461 nuclei and performs as well as MatchDown with both heads (\cref{tab:ablation,tab:bbbc038star}). Exact matching matters where cells are scarce: on BBBC039 at factors 16 to 32, rescue loses more nuclei than MatchDown-P on 33 to 99\% of the images (\cref{tab:rescuefactor}); at factor 16 its heads have 1.4 to 3.1 points higher \AccF, and only the affinity head counts worse, by 0.60 nuclei per image (\cref{tab:rescuesoft}).
Soft labels match or exceed MatchDown's count reduction on BBBC038 with cross-validated thresholds, at 8.2 \AccF points below majority.
At factor 16 their count error is 11.5 nuclei per image higher than MatchDown-P's at the default thresholds, because a nucleus that fills less than half of every cell it touches never reaches a foreground value of 0.5.
Training with a full-resolution loss, where one can be computed, gives the same count reduction as MatchDown with \AccF 2.9 points higher (\cref{tab:ablation}).

\section{Conclusion}
\label{sec:conclusion}
The regions a coarse target can keep are determined exactly by a matching problem: every region of a cell or more can be kept, a minimum-cost matching attains the maximum, and every unavoidable loss has a certificate.
Keeping a region in the target and reconstructing its shape are separate bottlenecks: the regions that majority pooling loses cannot be reconstructed above IoU 0.5 by upsampling any coarse target, yet on BBBC038 a star-convex head trained on MatchDown targets segments 35\% of them above that IoU, against 15\% when trained on majority targets and 24\% on StarDist's strided target.
Across the three star-convex settings, with cross-validated thresholds, this head has the lowest count error of the targets in \cref{tab:headline} and \AccF 2.6 to 4.2 points above the strided target.
An affinity head, with cell-level output, lowers count error but loses \AccF, because it can only reconstruct what the cells can.

\paragraph{Limitations.}
Training covers two nucleus datasets, factors 2 to 16 and two heads, without natural images; BBBC039 at factor 16 is a stress test.
Thresholds are cross-validated for count error, so \AccF and F1 are count-calibrated, and on BBBC038 the star-convex count ranking depends on the NMS range (\cref{tab:nms}).

\section*{LLM Usage Statement}
We used large language model assistants (Claude, Anthropic; Codex, OpenAI) to write, run and review the implementation, experiment, analysis and figure code, to check reported numbers against the run outputs, and to edit the text. The authors designed the method and the experiments, verified every proof, number and reference, and take full responsibility for the paper.

{
    \small
    \bibliographystyle{ieeenat_fullname}
    \bibliography{main}
}
\clearpage
\maketitlesupplementary
\appendix
\crefalias{section}{appendix}\crefalias{subsection}{subappendix}
\renewcommand{\thetable}{S\arabic{table}}
\renewcommand{\thefigure}{S\arabic{figure}}
\setcounter{table}{0}
\setcounter{figure}{0}
\renewcommand{\topfraction}{0.95}\renewcommand{\bottomfraction}{0.95}\renewcommand{\textfraction}{0.03}\renewcommand{\floatpagefraction}{0.8}
\setcounter{topnumber}{4}\setcounter{bottomnumber}{3}\setcounter{totalnumber}{6}
\makeatletter\setlength{\@fptop}{0pt}\setlength{\@dblfptop}{0pt}\makeatother
\noindent\Cref{sec:supp:proofs} proves the results of \cref{sec:method} and describes how we verified the implementation, \cref{sec:supp:data} defines the datasets, metrics and baselines, \cref{sec:supp:training} gives the training protocol, and \cref{sec:supp:audit,sec:supp:trainres} report the label-level and training results.

\section{Proofs}
\label{sec:supp:proofs}

\paragraph{Notation.}
Regions $r\in[R]$, cells $c\in[C]$ and counts $n_{rc}$ are as in \cref{sec:method:problem}.
Let $m_c=\max_r n_{rc}$ and $D_{\rm maj}=P-\sum_c m_c$.
Let $\ell(r,c)=m_c-n_{rc}\ge0$ on supported edges, so that every labeling satisfies $D(a)=D_{\rm maj}+\sum_c\ell(a(c),c)$.

\subsection{Proof of \texorpdfstring{\cref{thm:assignment}}{Theorem 3}}
\textbf{Step 1 (upper bound and attainability).}
Select one cell carrying each region represented in $a$.
The selected cells are distinct and their edges form a matching, so $K(a)\le\nu(G)$.
Conversely, label the cells of a maximum matching with their matched regions and fill the rest arbitrarily within their support.
At least $\nu(G)$ regions survive, and by the bound no more.

\textbf{Step 2 (penalty dominance).}
A full matching with $k$ real edges has weight $W=R+B(R-k)+\Lambda$, where $\Lambda=\sum\ell$ over its real edges.
Real-matched cells are distinct, so $0\le\Lambda\le\sum_c m_c\le P<B$.
Raising $k$ by one lowers $W$ by at least $B-P\ge1$.
A minimum-weight full matching therefore has $k=\nu(G)$ real edges, and among those it minimizes $\Lambda$.

\textbf{Step 3 (majority completion).}
Cells not used by the matching keep a majority region and contribute zero cost, so $D(a_*)=D_{\rm maj}+\Lambda_*$.
An unmatched region cannot be the majority region of such a cell: that would give $\nu(G){+}1$ distinct representatives, contradicting Step 1.

\textbf{Step 4 (pixel optimum).}
For any $a$ with $K(a)=\nu(G)$, choose distinct representative cells.
Their cost is a sub-sum of nonnegative cell costs, so $\Lambda_*\le\Lambda_a\le D(a)-D_{\rm maj}$.
Hence $D(a_*)\le D(a)$.
The implementation enforces $R(B+1)+P<2^{52}$, so that float64 weights are exact integers.
\hfill$\square$

\subsection{Algorithm, one-cell survival and a worked example}
\begin{algorithm}[t]
\caption{MatchDown-P: a coarse target with maximum survival}
\label{alg:matchdown}
\begin{algorithmic}[1]
\State $n_{rc}\gets$ the pixels of region $r$ (4-connected components of the labels) in cell $c$; $G$ has an edge wherever $n_{rc}>0$
\State $\tau(r)\gets 4,3,2,1$ for evaluated regions of at least $f^2$ pixels, smaller evaluated regions, non-evaluated regions of at least $f^2$ pixels, smaller non-evaluated regions
\State $B\gets P+1$; edge $(r,c)$ costs $1+\max_{r'}n_{r'c}-n_{rc}+B\,(4-\tau(r))$; each region also has a private dummy cell at cost $4B+1$
\State solve the minimum-cost full matching of the regions \cite{jonker1987shortest}
\State $a(c)\gets$ the majority region of $c$ for every cell, then $a(c)\gets r$ for every matched real edge $(r,c)$
\State repair: at most two sweeps of single-cell relabelings under the conditions of \cref{sec:method:repair}
\State certificate: the regions reachable by alternating paths from the unmatched regions, in clusters of $k_i$ regions sharing $m_i<k_i$ cells (\cref{thm:certificate})
\end{algorithmic}
\end{algorithm}
\begin{figure*}[t]
\centering
\includegraphics[width=0.78\textwidth]{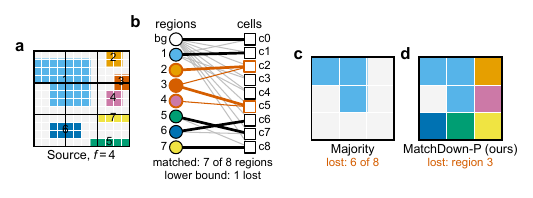}
\caption{\textbf{A worked example of survival as matching.} (a) A $12{\times}12$ source at $f{=}4$: seven regions on a background. (b) Its support graph. Bold edges are a maximum matching ($\nu{=}7$ of 8 regions); regions 2, 3 and 4 share only cells c2 and c5, so one of them must be lost, and this Dulmage--Mendelsohn cluster certifies the single unavoidable loss (\cref{thm:certificate}). (c) Majority pooling keeps 2 of the 8 regions, the background and region 1. (d) MatchDown-P keeps 7 of 8 and loses only region 3, one of the three regions of the cluster certified in (b); the matching drawn in (b) loses region 4 instead, at the same survival.}
\label{fig:matching}
\end{figure*}
\noindent\textbf{Proof of \cref{thm:onecell}.} Take $k$ regions of at least $f^2$ pixels each, and let $N$ be the set of cells they touch.
Those cells hold at least $k f^2$ of their pixels, and a cell holds at most $f^2$ pixels, so $|N|\ge k$.
Hall's condition therefore holds for the large regions, and a matching saturates them \cite{hall1935}.
By the augmentation property of the matroid of \cref{thm:matroid}, it extends to a survivable set of $\nu(G)$ regions that still contains them, which is the survivor set of a labeling with $K(a)=\nu(G)$. \hfill$\square$

\begin{proposition}[Tier optimality]
\label[proposition]{prop:tiers}
For every $t$, the MatchDown-P labeling keeps the largest number of regions of the $t$ highest tiers that any labeling can keep. Before the repair (line 6 of \cref{alg:matchdown}), it has the least pixel disagreement among the labelings that do so for every $t$; the repair keeps the surviving regions and adds at most $\lfloor P/200\rfloor$ to the disagreement.
\end{proposition}
\noindent\textbf{Proof.} Keeping a region of tier $\tau$ instead of its dummy saves $B\tau-\ell$ with $0\le\ell<B$ and the total of all $\ell$ below $B$, so a minimum-cost full matching maximizes $\sum_{r\ \mathrm{kept}}\tau(r)$ first and minimizes $\sum\ell$ second. On a matroid, the independent sets of maximum weight for a weight that is constant on each tier are exactly the bases that, for every $t$, contain the most elements of the $t$ highest tiers (Gale optimality of the greedy algorithm, \cref{thm:matroid}). Among these, the second criterion is the disagreement of Step 4 of the proof of \cref{thm:assignment}. \hfill$\square$

\subsection{Repair invariants}
Starting from $a_*$, a repair sweeps the cells at most twice and relabels cell $c$ from $u$ to $v$ only if $v$ is in the support of $c$ and already survives, $u$ keeps another cell, $D$ stays within $D(a_*)+\lfloor P/200\rfloor$, the pair $(F,L)$ strictly decreases lexicographically, and neither $u$ nor $v$ gains a 4-connected coarse component.
\begin{proposition}[Repair invariants]
\label[proposition]{prop:repair}
Every accepted relabeling of the repair of \cref{sec:method:repair} keeps the set of surviving regions and every cell supported, keeps $D\le D(a_*)+\lfloor P/200\rfloor$, strictly decreases $(F,L)$, and never increases any $b_r$; the number of relabelings is at most $\min\{2C,(|\mathcal S|+1)F(a_*)+L(a_*)\}$, with $\mathcal S$ the set of source adjacencies.
\end{proposition}
Consider an accepted relabeling of cell $c$ from $u$ to $v$.
\begin{enumerate}
\item \textbf{Survivors and support.}
$v$ is taken from the support of $c$ and from the surviving regions, and $u$ keeps another cell.
No region appears or disappears, and all other cells are unchanged.
\item \textbf{Budget.}
Only cell $c$ changes, so $D$ changes by exactly $n_{uc}-n_{vc}$, which is checked against the budget.
\item \textbf{Adjacency descent.}
A single-cell change affects at most four grid edges.
The multiplicities of adjacent region pairs are updated exactly, so $F$ and $L$ are exact.
Only candidates with $(F',L')<_{\rm lex}(F,L)$ are accepted.
\item \textbf{Component counts.}
The guard compares the 4-connected component counts of $u$ and $v$ on the whole grid before and after the change, and accepts only if neither increases.
Every other region's mask is unchanged.
By induction, $b_r(a_t)\le b_r(a_0)$ for all $r$.
\item \textbf{Termination.}
$\Phi=(|\mathcal S|+1)F+L$ drops by at least one per accepted relabeling, because a drop in $F$ can be offset by at most $|\mathcal S|$ in $L$.
At most one relabeling is accepted per cell per sweep, and there are two sweeps.
\end{enumerate}
\noindent\textbf{4-connectivity is essential.}
Take region pixels at $(1,0)$, $(1,1)$ and $(0,2)$ and remove $(1,1)$.
The 4-connected count stays at 2, while the 8-connected count rises from 1 to 2.
\hfill$\square$

\subsection{Forced adjacencies}
\begin{figure}[t]
\centering
\includegraphics[width=\columnwidth]{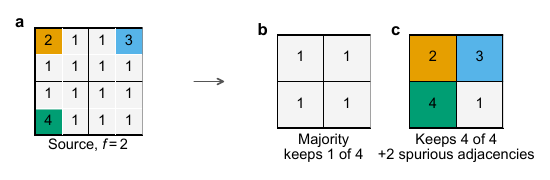}
\caption{\textbf{Spurious adjacencies forced by maximum survival.} (a) A $4{\times}4$ source at $f{=}2$: a connected background with single-pixel regions at three corners. (b) Majority pooling keeps one region and creates no spurious adjacency. (c) Of the eight supported $2{\times}2$ labelings, only this one keeps all four regions, and it creates the adjacencies $\{2,3\}$ and $\{2,4\}$, which the source does not have.}
\label{fig:forced}
\end{figure}
\begin{proposition}[Forced adjacencies]
\label[proposition]{prop:forced}
On the $4{\times}4$ source of \cref{fig:forced} at $f{=}2$, exactly one of the eight supported labelings keeps all four regions, and it has two spurious adjacencies; majority pooling keeps one region and has none.
\end{proposition}
The four cells of \cref{fig:forced} have supports $\{1,2\}$, $\{1,3\}$, $\{1,4\}$ and $\{1\}$, in row order.
This gives $2\cdot2\cdot2\cdot1=8$ supported labelings.
Only $\begin{psmallmatrix}2&3\\4&1\end{psmallmatrix}$ contains all four regions.
It has the coarse adjacencies $\{2,3\},\{2,4\},\{1,3\},\{1,4\}$.
Against the source adjacencies $\{1,2\},\{1,3\},\{1,4\}$, two are spurious and $\{1,2\}$ is missing.
\hfill$\square$

\subsection{Survival under a coverage threshold}
\label{sec:supp:kappatheory}
For $\kappa\in\{1/8,2/8,4/8\}$ let $G_\kappa$ keep the edges with $n_{rc}\ge\kappa|c|$.
Cells with no $G_\kappa$ edge (\emph{uncovered} cells) take their majority region, and a region is \emph{kept with coverage} $\kappa$ if it holds a cell through a $G_\kappa$ edge; a labeling that obeys these rules is \emph{admissible}.
\begin{theorem}[Exact survival under coverage]
\label{thm:kappa}
The number of regions kept with coverage $\kappa$ by any admissible labeling is at most $\nu(G_\kappa)$, and the construction of \cref{thm:assignment} on $G_\kappa$ with fixed uncovered cells attains it with the least disagreement.
Let $R_0$ be the regions kept by uncovered cells.
Every admissible labeling loses at least as many evaluated regions as the deficiency of the evaluated regions outside $R_0$ in $G_\kappa-R_0$, and the tier-weighted construction in which every region of $R_0$ ranks below all other regions attains that minimum.
\end{theorem}
\noindent
The ranking of $R_0$ matters: at $f{=}4$ and $\kappa{=}1/4$, let one cell hold 9 pixels of $r$ and 7 of $s$, and its neighbor 3 pixels each of $r,a,b,c$ and 2 each of $d,e$, with $r$ first in index order and $r$ 4-connected across both cells.
The neighbor is uncovered and takes $r$; a matching that ranks $r$ and $s$ equally gives the first cell to $r$, at lower pixel cost, and loses $s$, whereas ranking $r$ last keeps both.
Our implementation ranks $R_0$ last, and on every image of \cref{tab:kappa} its loss equals the minimum computed separately as in the proof.
On Cityscapes at $\kappa{=}1/8$ the minimum, as a fraction of the losses of class-first majority pooling, is 0.5621; the unweighted construction reaches 0.5627.

\noindent\textbf{Proof.}
Let $C^+$ be the cells incident to $G_\kappa$, and $C^0$ the uncovered cells, fixed to their lowest-index majority region.
\begin{itemize}
\item \textbf{Upper bound.}
The witnessing cells of distinct regions kept with coverage $\kappa$ are distinct cells of $C^+$ joined by $G_\kappa$ edges, so they form a matching of $G_\kappa$.
\item \textbf{Attainability.}
Label a maximum matching of $G_\kappa$, fill the other $C^+$ cells with any region joined to them in $G_\kappa$, and keep $C^0$ fixed.
\item \textbf{Pixel optimum.}
Steps 2--4 of the proof of \cref{thm:assignment} apply to $C^+$, with dummy penalty $P_\kappa+1$ where $P_\kappa$ is the number of pixels on $G_\kappa$ edges; $C^0$ contributes a fixed disagreement.
\item \textbf{Minimum evaluated loss.}
Regions of $R_0$ are kept whatever the covered cells hold.
The other regions a labeling keeps with coverage hold distinct cells of $C^+$ through $G_\kappa$ edges, a matching of $G_\kappa-R_0$, so the evaluated ones among them number at most the largest number of evaluated regions that a matching of $G_\kappa-R_0$ saturates; regions with no $G_\kappa$ edge outside $R_0$ are lost.
With $R_0$ ranked last, the tier weights make the minimum-cost matching saturate that largest number (\cref{prop:tiers}, with the regions of $R_0$ in the lowest tiers).
\item \textbf{$\kappa{=}0$.}
$G_\kappa$ is the full support graph and $C^0=\varnothing$, so the construction is that of \cref{thm:assignment}.
\end{itemize}
The restricted repair adds three conditions to \cref{prop:repair}: both edges of a relabeling lie in $G_\kappa$, the old region keeps another cell with coverage $\kappa$, and the new region already has one.
Uncovered cells never change, so both survivor sets are unchanged and the component argument carries over.
Regions kept only by uncovered cells cannot be counted as kept with coverage $\kappa$: a $4{\times}4$ grid of 16 distinct pixels at $f{=}2$ and $\kappa{=}4/8$ has $\nu(G_\kappa)=0$ and four regions kept by uncovered cells.
\hfill$\square$

\paragraph{Lower bound from uncovered regions.}
Let $I_\kappa$ be the regions with no $G_\kappa$ edge, and $M_\kappa\subseteq I_\kappa$ those that are the fixed majority region of some uncovered cell.
In every admissible labeling a covered cell holds a region joined to it in $G_\kappa$, so a region of $I_\kappa$ can appear only in uncovered cells, which are fixed.
Hence every region of $I_\kappa\setminus M_\kappa$ is lost by every admissible labeling, and the evaluated ones give the lower bound of \cref{tab:kappa}.
On Cityscapes and COCO at $\kappa{=}1/8$, $M_\kappa$ contained no evaluated region.
The losses of a $\kappa$-restricted labeling are these regions plus the evaluated regions with $G_\kappa$ edges that the restricted matching cannot place.

\subsection{Rescue bounds}
\label{sec:supp:gadget}
\begin{proposition}[Rescue bounds]
\label[proposition]{prop:rescue}
A rescue that gives each lost region a cell of its support not needed by a surviving region keeps at least $\nu(G)/2$ regions.
Greedy rescue, which takes each lost region's best-covered cell, has no such bound: for every $k$ there is a map at $f{=}4$ with $\nu(G)=4k+1$ on which it keeps $k+1$ regions, while the matching keeps all $4k+1$.
\end{proposition}
\noindent
The first statement holds because the representative cells of the surviving regions form a maximal matching, which has at least half the edges of a maximum one.
For the second, at $f{=}4$ block $i$ is a $12{\times}8$ array of $3{\times}2$ cells on a common background $t$.
Cell $A$ holds 10 pixels of $s_i$ and 2 pixels of each of $l_{i1},l_{i2},l_{i3}$, and each $l_{ij}$ continues with one pixel into its own neighboring cell $C_j$, whose other 15 pixels are $t$; the two remaining cells are all $t$.
Majority pooling keeps $s_i$ and $t$.
The best-covered cell of each $l_{ij}$ is $A$ (2 pixels against 1), which holds $s_i$ and would remove it, so greedy rescue keeps $k{+}1$ regions of $\nu(G)=4k{+}1$, while the matching and maximal rescue keep all $4k{+}1$.
\hfill$\square$

\subsection{Verification of the implementation}
\label{sec:supp:checks}
We verified the implementation against exhaustive enumeration of every supported labeling on small grids (survivor count, disagreement, the coverage construction and the per-region IoU bound), against a Hopcroft--Karp maximum matching written independently of the solver, and against the invariants of \cref{prop:repair}, on synthetic label maps of seven kinds (random labels, blobs, thin lines, ties, clipped borders, disconnected labels and one region per pixel) at factors 2 to 16.
All checks passed; the tests will be released with the code.

\section{Datasets, metrics and baselines}
\label{sec:supp:data}
\begin{table*}[t]
\centering\footnotesize
\caption{\textbf{Datasets.} Evaluated regions are the 4-connected components of instances and of stuff classes that a metric scores (not void or crowd); for the nucleus datasets the second number counts annotated nuclei, of which a few consist of several 4-connected pieces. COCO-A, COCO-B and COCO-R are disjoint random sets of COCO panoptic validation images; COCO-R was held out from all development. Majority loss: share of evaluated regions (COCO, Cityscapes) or nuclei (BBBC038, BBBC039) missing from the majority-pooling target at factor 8 ($^\dagger$factor 16). $^\ddagger$From $256{\times}256$ to $1040{\times}1388$ pixels. Use: L, label-level evaluation (factors 2 to 16 unless stated); T, training; BSDS500 serves the repair-optimality and binary-topology comparisons (\cref{sec:supp:labelcmp}).}
\label{tab:datasets}
\setlength{\tabcolsep}{3pt}
\begin{tabular}{@{}llrrrrl@{}}
\toprule
Dataset & Image size & Images & Evaluated regions & Per image & Majority loss (\%) & Use \\
\midrule
COCO-A & up to $640{\times}640$ & 1,000 & 25,860 & 25.9 & 34.9 & L \\
COCO-B & up to $640{\times}640$ & 1,000 & 25,909 & 25.9 & 34.7 & L \\
COCO-R & up to $640{\times}640$ & 1,000 & 25,963 & 26.0 & 35.5 & L \\
Cityscapes (6 cities) & $1024{\times}2048$ & 900 & 104,234 & 115.8 & 27.9 & L \\
BBBC038 (stage 1) & varied$^\ddagger$ & 670 & 29,668 / 29,461 & 44.3 & 8.8 & L, T ($f{=}8$) \\
BBBC039 & $520{\times}696$ & 197 & 23,615 nuclei & 119.9 & 10.1$^\dagger$ & T, L ($f{=}8$--32) \\
BSDS500 & $481{\times}321$ & 200 crops of $64{\times}64$; 50 masks & -- & -- & -- & repair, binary \\
\bottomrule
\end{tabular}
\end{table*}

\paragraph{Datasets.}
\Cref{tab:datasets} summarizes the datasets.
We chose the second training dataset among BBBC039 and NuInsSeg \cite{mahbod2024nuinsseg} at factors 8 and 16 by a rule fixed before training: among the pairs whose majority and MatchDown-P targets differ by at least 2 nuclei per image in oracle count error, take the largest difference.
The differences are 11.7 (BBBC039, factor 16), 6.9 (BBBC039, factor 8), 2.6 (NuInsSeg, factor 16) and $-0.2$ (NuInsSeg, factor 8); BBBC038 at factor 8 gives 3.7.
BBBC039 at factor 16 is therefore a stress test with many lost nuclei (10.1\% under majority pooling).
We use the 197 of its 200 fields that contain nuclei.
A small fraction of BBBC039 images also appear in BBBC038; each dataset is split, trained and evaluated on its own, so no image is used for training and evaluation of the same model.

\paragraph{Metrics.}
\label{sec:supp:metrics}
Count error is the absolute difference between the number of predicted objects and the number of nuclei in an image.
At a fixed decoder operating point, $\mathrm{Acc}_\tau=\mathrm{TP}/(\mathrm{TP}+\mathrm{FP}+\mathrm{FN})$ with one-to-one matching at IoU${}>\tau$; \AccF is $\mathrm{Acc}_{0.5}$ and \Acc the mean over $\tau=0.50,0.55,\dots,0.95$.
These scores are computed at one operating point, without integrating over a precision--recall curve as COCO AP does.
FP50 is the number of false positives per image at IoU${}>0.5$.
Centroid F1 matches predicted and true objects one to one by increasing centroid distance, up to 8 pixels.
F1 at IoU${}>0.5$ is $2\mathrm{TP}/(2\mathrm{TP}+\mathrm{FP}+\mathrm{FN})$ with the counts of \AccF pooled over images and seeds, so it weights large images more than \AccF, which averages per-image values.
A nucleus missing from a target counts as \emph{detected} when a greedy one-to-one matching by overlap gives it its own predicted object that is no larger, in cells, than its support plus one cell, and as \emph{matched} when that object reaches IoU${}>0.5$.
PQ is averaged over classes, $\mathrm{PQ}=\frac1{|K|}\sum_{k}\mathrm{SQ}_k\,\mathrm{RQ}_k$, with $\mathrm{SQ}_k$ the mean IoU of the matched segments of class $k$ and $\mathrm{RQ}_k=\mathrm{TP}_k/(\mathrm{TP}_k+\tfrac12\mathrm{FP}_k+\tfrac12\mathrm{FN}_k)$; a PQ difference splits exactly into an SQ term, an RQ term and their product, computed per class from the pooled counts and averaged over classes.

\paragraph{Downsampling methods.}
\emph{Strided} sampling takes the first pixel of each cell (\texttt{lbl[::f,::f]}, PyTorch's default nearest interpolation and StarDist's target).
\emph{Nearest (cell center)} takes pixel $\lfloor(f-1)/2\rfloor$ of each cell along each axis, and \emph{nearest-exact} pixel $f/2$ (PyTorch's nearest-exact, used by nnU-Net's deep supervision).
\emph{Majority pooling} gives each cell its most frequent region (lowest index on ties); the \emph{class-first} variant takes the most frequent class, then the most frequent region of that class, with void and crowd counted as classes.
\emph{Max pooling} takes the largest label index in each cell.
\emph{One-hot bilinear + argmax} interpolates the one-hot label maps bilinearly and takes the argmax.
\emph{nnU-Net per-label} interpolation is \texttt{resize\_segmentation} of batchgenerators with order 1: each region is interpolated separately and written where it reaches 0.5, in ascending order, and our implementation reproduces the library output exactly.
\emph{Smallest-region priority} gives each cell the smallest region that covers at least a quarter of it.
\emph{Greedy} and \emph{maximal rescue} are defined in \cref{sec:method:rescue}, and \emph{dilated} targets label every cell a nucleus touches with that nucleus.

\FloatBarrier
\section{Training details}
\label{sec:supp:training}
\begin{table*}[t]
\centering\footnotesize
\caption{\textbf{Training and decoding settings.} Every model of a seed and fold uses the same initialization, image crops and augmentations for all targets, so that differences between targets are paired.}
\label{tab:hyper}
\setlength{\tabcolsep}{5pt}
\begin{tabular}{@{}p{0.17\textwidth}p{0.37\textwidth}p{0.38\textwidth}@{}}
\toprule
 & Affinity head & Star-convex head \\
\midrule
Backbone & \multicolumn{2}{l}{ResNet-18 (ImageNet weights) with a feature pyramid of 128 channels, laterals down to the target stride} \\
Output stride & 8 (BBBC038), 16 (BBBC039) & 8 (BBBC038, also 4 and 2), 16 and 8 (BBBC039) \\
Outputs per cell & foreground, affinities to 4 of the 8 neighbors & object probability, center offset (2), 32 ray distances \\
Loss & BCE on foreground over all cells + BCE on affinities over valid cell pairs & BCE on probability + $0.2\times$L1 on offsets + $0.2\times$L1 on distances (assigned cells) \\
Optimizer & \multicolumn{2}{l}{AdamW, learning rate $5{\times}10^{-4}$, $\beta=(0.9, 0.999)$, weight decay $10^{-4}$} \\
Schedule & \multicolumn{2}{l}{500-iteration linear warm-up, then polynomial decay (power 0.9); 8,000 iterations} \\
Batch and crops & \multicolumn{2}{l}{16 crops of $256{\times}256$ pixels at offsets that are multiples of 16, so targets of every method align} \\
Augmentation & \multicolumn{2}{l}{one of the 8 rotations and flips by quarter turns, applied exactly to image and labels} \\
Input & \multicolumn{2}{l}{ImageNet normalization; 16-bit BBBC039 images stretched to 8 bits between the 1st and 99.8th percentiles} \\
Precision, hardware & \multicolumn{2}{l}{bfloat16; one NVIDIA A100 (BBBC038) or RTX 4090 (BBBC039) per run} \\
Seeds and folds & \multicolumn{2}{l}{9 seeds $\times$ 3 folds; each image is predicted by the model of the fold that did not train on it} \\
Decoding & foreground $>t_{\mathrm{fg}}$; neighbors joined if affinity $>t_{\mathrm{aff}}$; connected components & candidates with probability $>t_p$, greedy NMS at polygon IoU $t_{\mathrm{nms}}$ \\
Default thresholds & $t_{\mathrm{fg}}=t_{\mathrm{aff}}=0.5$ & $t_p=0.5$, $t_{\mathrm{nms}}=0.4$ (StarDist) \\
Cross-validation grid & $t_{\mathrm{fg}},t_{\mathrm{aff}}\in\{0.05,0.10,\dots,0.95\}$ & $t_p\in\{0.1,\dots,0.9\}$; $t_{\mathrm{nms}}\in\{0.3,0.4,0.5\}$ (BBBC038), $\{0.1,\dots,0.4\}$ (BBBC039) \\
\bottomrule
\end{tabular}
\end{table*}

\paragraph{Folds and targets.}
The 670 BBBC038 stage-1 training images are split into three folds stratified by image size, and the 197 BBBC039 images into three folds; each fold's model trains on the other two, so every image is predicted once per target and seed by a model that never saw it.
Every target is computed once per image before training and cropped and transformed with the image.
On BBBC038, the nearest-target affinity model in \cref{tab:training}, all star-convex models (\cref{tab:bbbc038star,tab:nms}), and the models in \cref{tab:controls} use a second run set with the same protocol. The cell-center affinity model in \cref{tab:headline} was rerun in the first set. Every difference is computed within one run set.

\paragraph{Targets of the two heads.}
The affinity head's target gives each cell its region's foreground label and each pair of neighboring foreground cells an affinity of 1 when they hold the same region; unassigned cells are background.
Soft labels replace these by the nucleus fraction of each cell and, for two neighboring cells, the probability that a nucleus pixel drawn from each belongs to the same nucleus \cite{alcover2026soft}.
The full-resolution loss upsamples the stride-$f$ logits bilinearly and supervises every pixel with the foreground and the same-nucleus label of the pixel pair that its cell-level channel describes.
In the masked target, the cells of the nuclei missing from the majority target carry no loss.
Every cell of the star-convex target that is assigned to a nucleus is supervised with the same description of it: probability 1, the offset from the cell center to the pixel of the nucleus farthest from its boundary (which always lies inside the nucleus), and the 32 ray distances from that pixel, computed as in StarDist.

\paragraph{Threshold cross-validation.}
For each model, seed and held-out fold, the cross-validated thresholds minimize the mean count error over the images of the other two folds on the grid of \cref{tab:hyper}, each image scored by its own held-out model; ties go to the point nearest the default, then to the smaller thresholds.
For the equal-FP point, each model takes the pair with the lowest count error among those whose FP50 on the tuning folds does not exceed the majority model's at its cross-validated pair, or the pair with the fewest false positives if none does.
With every point of the grid decoded, the default points reproduce the default-threshold scores exactly.
Because the thresholds minimize count error, \AccF and F1 at them describe a count-calibrated operating point.
The models that predict the tuning folds were trained on images that include the evaluated fold.
In a fully nested check, we selected the affinity thresholds of BBBC038 on one half of each held-out fold and evaluated them on the other half, so that neither the model nor the thresholds saw the evaluated images.
The MatchDown model's count error is then 1.03 nuclei per image lower than the majority model's, lower in all nine seeds, with \AccF 1.2 points lower and 1.6 more false positives per image, the same pattern as with cross-validated thresholds; the equal-FP constraint can be met in 37 of the 54 seed, fold and half combinations.

\paragraph{Confidence intervals.}
A difference between two targets is the mean over the nine seeds of the per-seed mean over images of the per-image difference.
Its 95\% interval comes from 10,000 bootstrap samples that resample seeds and then images, with all targets sharing each resample, so that the pairing of targets on the same seed, fold and image is kept \cite{efron1979bootstrap,diciccio1996bootstrap}; the seed-level spread is scaled by a Student-$t$ factor for nine seeds.
Thresholds are fixed across bootstrap samples, so the intervals cover the variation over seeds and images.

\FloatBarrier
\section{Label-level results}
\label{sec:supp:audit}
\begin{table*}[t]
\centering\footnotesize
\caption{\textbf{Evaluated regions lost (\%) by every method, dataset and factor.} Lower bound: the minimum of \cref{thm:certificate}; reachable: share of evaluated regions whose IoU bound exceeds 0.5 (\cref{prop:ceiling}). MatchDown-P equals the lower bound in every column. At $f{=}2$ the cell-center pixel is the first pixel of the cell, so nearest (cell center) equals strided. Majority pooling counts the most frequent region in each cell; the class-first variant picks the most frequent class, then its most frequent region. On COCO-A and COCO-B every entry is within one percentage point of COCO-R. Best in bold, second best underlined.}
\label{tab:supp:lost}
\setlength{\tabcolsep}{3.6pt}
\begin{tabular}{@{}l*{12}{r}@{}}
\toprule
 & \multicolumn{4}{c}{COCO-R (25,963 regions)} & \multicolumn{4}{c}{Cityscapes (104,234 regions)} & \multicolumn{4}{c}{BBBC038 (29,668 regions)} \\
\cmidrule(lr){2-5}\cmidrule(lr){6-9}\cmidrule(lr){10-13}
Method & $f{=}2$ & 4 & 8 & 16 & $f{=}2$ & 4 & 8 & 16 & $f{=}2$ & 4 & 8 & 16 \\
\midrule
Strided (StarDist) & 17.6 & 26.5 & 32.7 & 40.1 & 8.7 & 15.1 & 24.0 & 38.3 & 0.42 & 0.95 & 6.2 & 26.4 \\
Nearest (cell center) & 17.6 & 26.8 & 32.8 & 40.2 & 8.7 & 15.0 & 24.0 & 38.1 & 0.42 & 0.69 & 5.3 & 26.2 \\
Nearest-exact (nnU-Net) & 17.9 & 26.9 & 32.7 & 40.1 & 8.6 & 15.0 & 24.0 & 38.1 & 0.41 & 0.78 & 5.2 & 26.0 \\
Max pooling & 12.9 & 21.4 & 30.2 & 41.6 & 9.8 & 17.5 & 28.2 & 41.6 & 0.23 & 0.43 & 1.3 & 7.4 \\
Majority pooling & 24.2 & 30.5 & 35.5 & 43.2 & 14.2 & 18.7 & 27.9 & 44.0 & 0.58 & 1.1 & 9.3 & 39.0 \\
Majority, class first & 24.0 & 30.7 & 35.5 & 43.1 & 10.7 & 17.9 & 27.6 & 43.9 & 0.51 & 1.1 & 9.3 & 39.0 \\
Greedy rescue & 4.7 & 10.8 & 19.0 & 29.6 & 0.24 & 1.2 & 3.4 & 10.8 & 0.04 & 0.11 & 0.52 & 4.8 \\
Maximal rescue & 4.2 & 9.7 & 16.9 & 26.5 & 0.19 & 0.88 & 2.5 & 7.1 & \underline{0.03} & \underline{0.10} & 0.20 & 2.1 \\
MatchDown & \underline{1.3} & \underline{6.2} & \underline{12.3} & \underline{18.7} & \underline{0.18} & \underline{0.82} & \underline{2.1} & \underline{5.1} & \textbf{0.02} & \underline{0.10} & \underline{0.14} & \underline{0.94} \\
MatchDown-P (ours) & \textbf{0.84} & \textbf{3.5} & \textbf{7.1} & \textbf{11.6} & \textbf{0.17} & \textbf{0.80} & \textbf{2.0} & \textbf{4.7} & \textbf{0.02} & \textbf{0.08} & \textbf{0.13} & \textbf{0.90} \\
\midrule
Lower bound (\cref{thm:certificate}) & 0.84 & 3.5 & 7.1 & 11.6 & 0.17 & 0.80 & 2.0 & 4.7 & 0.02 & 0.08 & 0.13 & 0.90 \\
Reachable (IoU bound $>0.5$) & 71.4 & 66.0 & 59.4 & 47.9 & 84.5 & 76.3 & 60.7 & 40.4 & 99.4 & 98.0 & 80.5 & 44.1 \\
\bottomrule
\end{tabular}
\end{table*}

\begin{table*}[t]
\centering\footnotesize
\caption{\textbf{Label-level evaluation at factor 8, further columns.} Lost (\%): evaluated regions lost on the two COCO development sets. $\ge$1 cell: lost regions of at least $f^2$ pixels per 1,000 evaluated regions. Recon.: share of the reachable regions that the target reconstructs at IoU${}>0.5$ under nearest upsampling. Void/crowd lost: share of the 157,318 void and crowd components of COCO-R missing from the target; MatchDown and MatchDown-P lose the same total number of components (70,559) and differ only in which ones. --: not computed for this method. Best in bold, second best underlined.}
\label{tab:supp:f8}
\setlength{\tabcolsep}{4pt}
\begin{tabular}{@{}lrrrrrrrrr@{}}
\toprule
 & \multicolumn{2}{c}{Lost (\%) $\downarrow$} & \multicolumn{3}{c}{$\ge$1 cell (per 1,000) $\downarrow$} & \multicolumn{3}{c}{Recon. (\%) $\uparrow$} & Void/crowd \\
\cmidrule(lr){2-3}\cmidrule(lr){4-6}\cmidrule(lr){7-9}
Method & COCO-A & COCO-B & COCO-R & Cityscapes & BBBC038 & COCO-R & Cityscapes & BBBC038 & lost (\%) $\downarrow$ \\
\midrule
Strided (StarDist) & 32.2 & 32.1 & 6.8 & 26.6 & 12.2 & 81.1 & 69.1 & 59.8 & 93.3 \\
Nearest (cell center) & 32.2 & 32.1 & 6.2 & 27.7 & 4.4 & 97.8 & 96.0 & \textbf{96.7} & -- \\
Nearest-exact (nnU-Net) & 32.3 & 32.0 & 5.8 & 27.4 & 5.3 & 97.7 & 95.9 & \textbf{96.7} & 93.3 \\
Max pooling & 29.5 & 30.0 & 35.9 & 76.8 & 1.6 & 74.5 & 64.6 & 60.3 & 97.8 \\
Majority pooling & 34.9 & 34.7 & 7.5 & 35.1 & 16.5 & \textbf{98.4} & \underline{96.1} & \underline{94.8} & 96.8 \\
Majority, class first & 34.9 & 34.6 & 8.0 & 33.0 & 16.5 & \underline{98.3} & \textbf{96.4} & \underline{94.8} & -- \\
Greedy rescue & 18.0 & 18.0 & 2.7 & 0.91 & \underline{0.34} & 89.9 & 94.8 & 94.1 & 44.1 \\
Maximal rescue & 16.2 & 16.1 & \underline{0.35} & \underline{0.02} & \textbf{0.00} & 89.6 & 94.7 & 94.1 & -- \\
MatchDown & \underline{11.9} & \underline{12.1} & \textbf{0.00} & \textbf{0.00} & \textbf{0.00} & 89.5 & 94.9 & 94.3 & \textbf{42.8} \\
MatchDown-P (ours) & \textbf{6.7} & \textbf{6.6} & \textbf{0.00} & \textbf{0.00} & \textbf{0.00} & 89.5 & 94.9 & 94.3 & \underline{43.7} \\
\bottomrule
\end{tabular}
\end{table*}

\begin{figure*}[t]
\centering
\includegraphics[width=\textwidth]{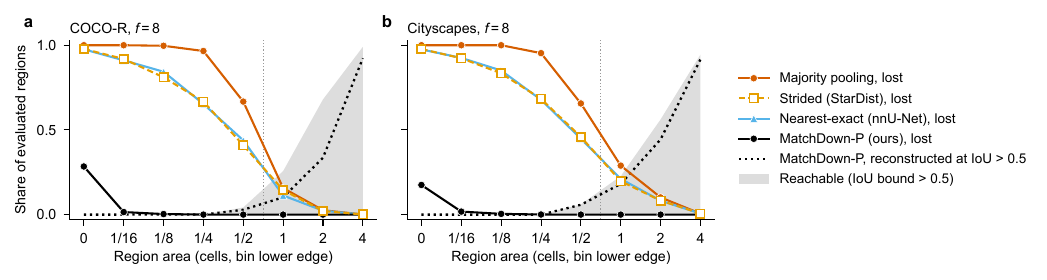}
\caption{\textbf{Loss by region area on COCO-R and Cityscapes at factor 8} (BBBC038 is \cref{fig:audit}b). Per area bin, labeled by its lower edge: the share of evaluated regions each method loses, the share that is reachable (IoU bound above 0.5, shaded), and the share that the MatchDown-P target reconstructs at IoU${}>0.5$. The dotted vertical line marks one cell.}
\label{fig:auditsupp}
\end{figure*}

\Cref{tab:supp:lost} gives the share of evaluated regions each method loses on COCO-R, Cityscapes and BBBC038 at factors 2 to 16, and \cref{tab:supp:f8} the remaining columns at factor 8; \cref{fig:auditsupp} shows the losses at factor 8 by region area on COCO-R and Cityscapes.
MatchDown-P's losses equal the lower bound of \cref{thm:certificate} for all 20 dataset--factor pairs (COCO-A, COCO-B, COCO-R, Cityscapes and BBBC038 at factors 2, 4, 8 and 16), and neither form of MatchDown loses an evaluated region of a cell or more on any of them.
Every other method does on every pair, except maximal rescue on nine pairs and greedy rescue on BBBC038 at factors 2 and 4.
On Cityscapes, MatchDown-P also equals the lower bound in each of the six cities.
Against MatchDown, the tiers lower the losses most on COCO (from 3,073 to 1,728 regions on COCO-A at factor 8) at a disagreement cost of at most 0.02\% of pixels; on the BBBC038 images both keep the same nuclei.
Keeping more regions adds spurious adjacencies: on COCO-A at factor 8, MatchDown-P has 4,634 spurious and 8,655 missing adjacencies among evaluated regions, against 3,367 and 10,196 for MatchDown.

\begin{figure*}[t]
\centering
\includegraphics[width=\textwidth]{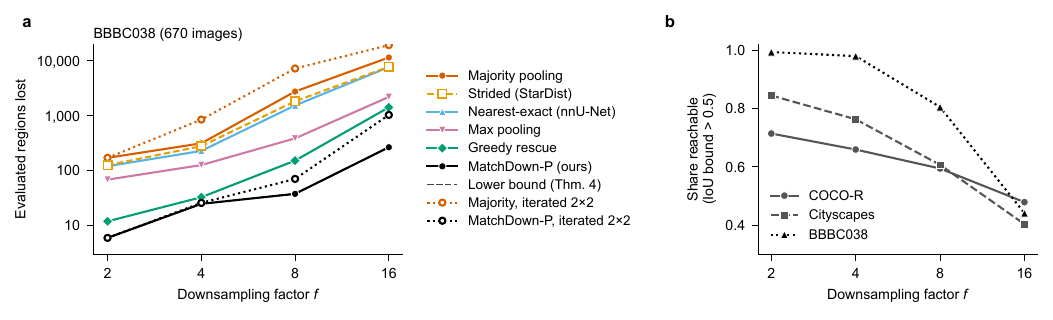}
\caption{\textbf{Every factor.} (a) Evaluated regions lost on BBBC038 at factors 2 to 16, with the label pyramids built by repeated $2{\times}2$ downsampling (dotted, open markers). Iterated majority pooling loses more than direct pooling to the same factor, and iterated MatchDown-P never loses fewer regions than direct MatchDown-P, which equals the lower bound at every factor. (b) Share of evaluated regions that are reachable (IoU bound above 0.5) on COCO-R, Cityscapes and BBBC038.}
\label{fig:factorsupp}
\end{figure*}
\begin{figure*}[t]
\centering
\includegraphics[width=\textwidth]{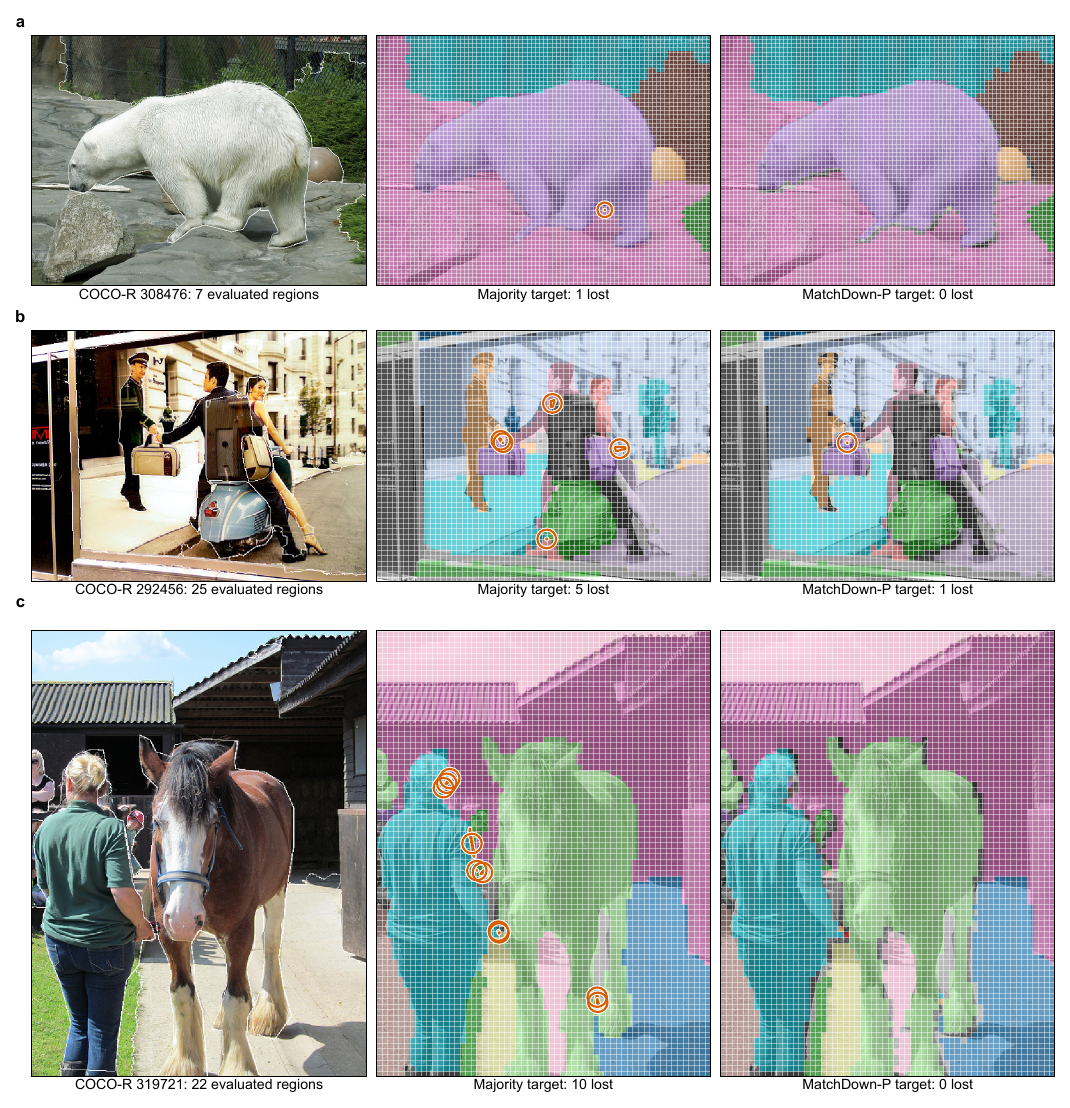}
\caption{\textbf{Majority and MatchDown-P targets at factor 8 on three COCO-R images}, at ranks 250, 500 and 749 of 1,000 by the per-image difference in lost evaluated regions. Left: the photograph with the region boundaries; middle and right: the majority-pooling and MatchDown-P targets drawn over it, void and crowd components in gray, each lost evaluated region outlined and ringed. (a) Image 308476, 7 evaluated regions, majority loses 1. (b) 292456, 25 regions, 5 against 1. (c) 319721, 22 regions, 10 against 0.}
\label{fig:panels:cocor}
\end{figure*}
\begin{figure*}[t]
\centering
\includegraphics[width=\textwidth]{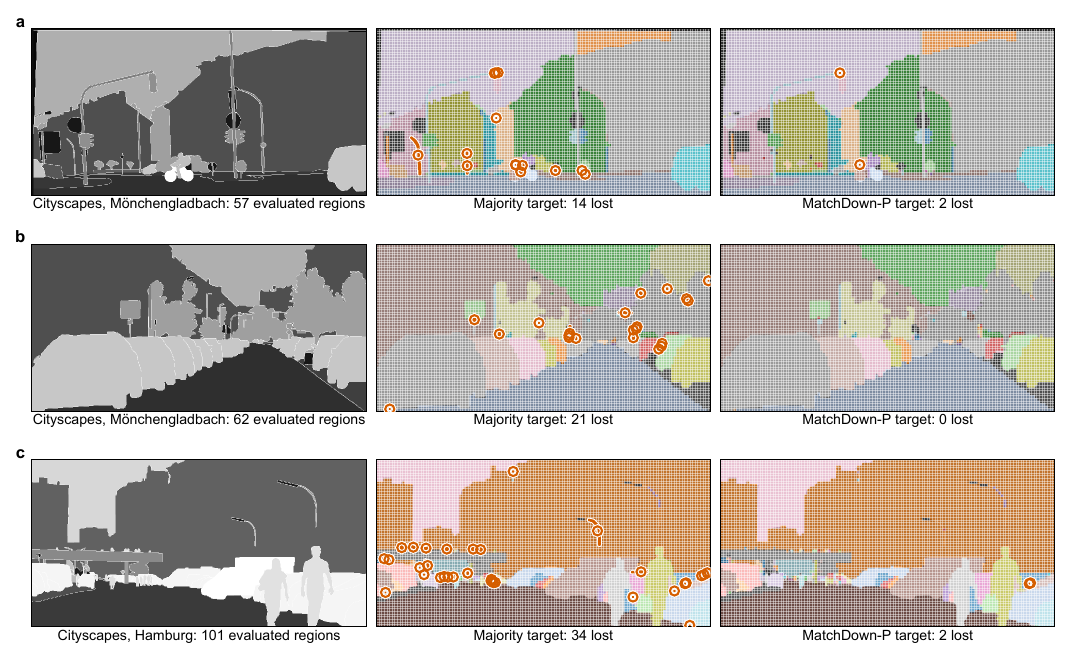}
\caption{\textbf{Majority and MatchDown-P targets at factor 8 on three Cityscapes frames}, chosen as in \cref{fig:panels:cocor} (ranks 225, 450 and 674 of 900). Each row shows the $512{\times}1024$-pixel window with the most regions lost by majority pooling; the left panel shows the semantic label map with the region boundaries. (a) M\"onchengladbach 000000\_000383, 57 evaluated regions in the window, majority loses 14 and MatchDown-P 2. (b) M\"onchengladbach 000000\_021104, 62 regions, 21 against 0. (c) Hamburg 000000\_105464, 101 regions, 34 against 2.}
\label{fig:panels:city}
\end{figure*}
\begin{figure*}[t]
\centering
\includegraphics[width=\textwidth]{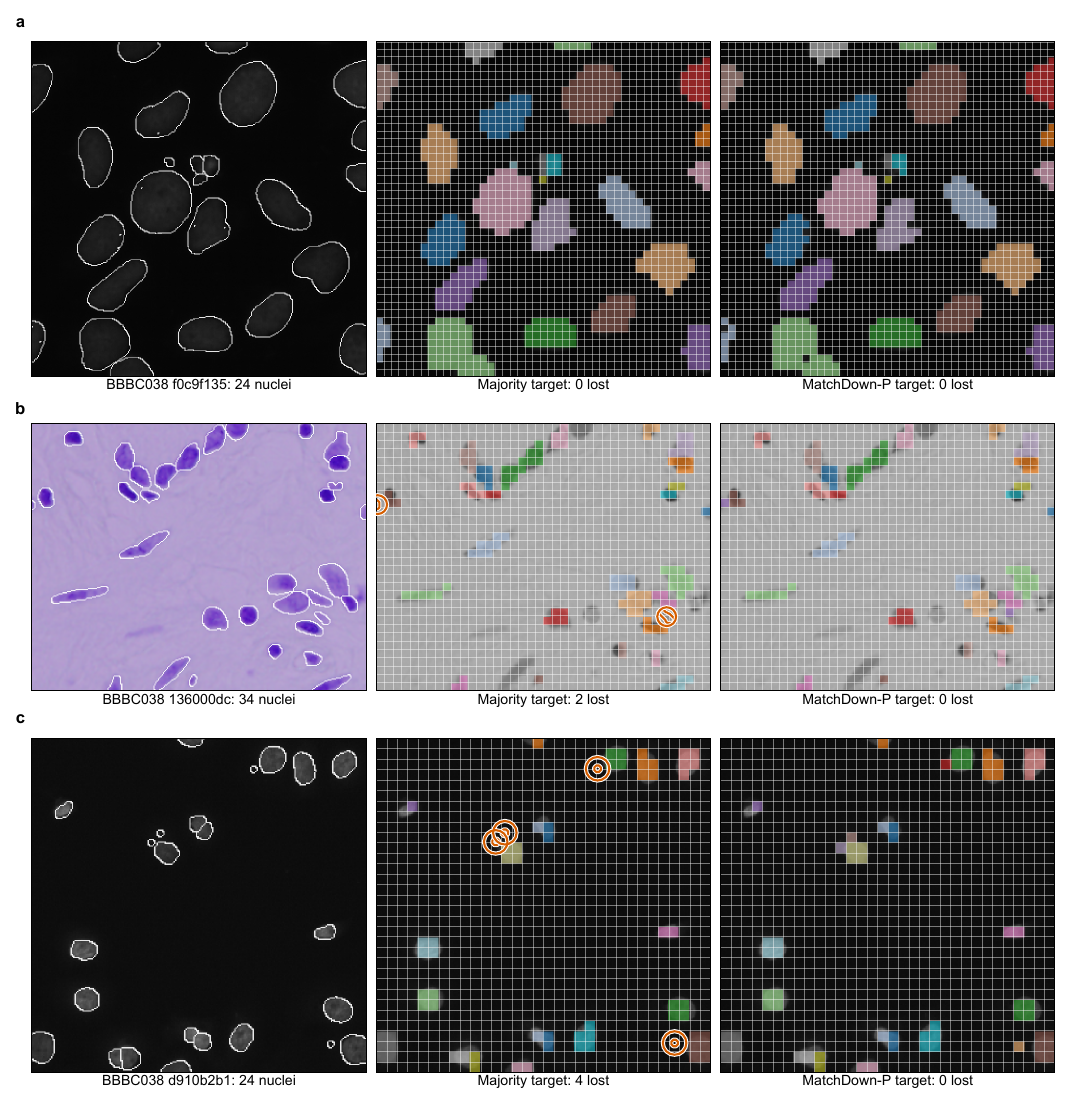}
\caption{\textbf{Majority and MatchDown-P targets at factor 8 on three BBBC038 images}, chosen as in \cref{fig:panels:cocor} (ranks 167, 334 and 502 of 670). (a) f0c9f135, 24 nuclei; neither target loses one, as on 27\% of the images. (b) 136000dc, 34 nuclei, majority loses 2 and MatchDown-P 0. (c) d910b2b1, 24 nuclei, 4 against 0.}
\label{fig:panels:nuc}
\end{figure*}

\Cref{fig:panels:cocor,fig:panels:city,fig:panels:nuc} show majority and MatchDown-P targets at factor 8 on images at the 25th, 50th and 75th percentiles of the per-image difference between the evaluated regions the two targets lose.

\paragraph{Label pyramids.}
A label pyramid built by repeated $2{\times}2$ majority pooling loses more than one direct pooling to the same factor: 7,277 evaluated regions against 2,772 on BBBC038 at factor 8 (\cref{fig:factorsupp}a).
Direct MatchDown-P never loses more than an iterated construction, because an iterated labeling is one supported labeling of the same grid.
On Cityscapes the pyramid loses 36,331 evaluated regions at factor 8 against 29,100 for direct pooling, and 9,302 against 9,207 on COCO-R; iterated MatchDown-P loses 71 against 38 on BBBC038 and 2,858 against 2,077 on Cityscapes.

\begin{table}[t]
\centering\footnotesize
\caption{\textbf{Reconstruction quality and survival on COCO-R at factor 2} (1,000 images, 25,963 evaluated regions): reconstruction PQ over all 133 classes, evaluated regions lost, and spurious adjacencies among evaluated regions per image. ``Void last'' and ``highest id'' change only how ties are broken; the tie rule alone moves the PQ of majority pooling by 1.08 to 1.15 points. MatchDown's PQ is 0.58 [0.48, 0.68] points (90\% interval) below class-first majority. At $f{=}2$, $\kappa{=}1/8$ and $2/8$ give the MatchDown target. Best in bold, second best underlined.}
\label{tab:cocor2}
\setlength{\tabcolsep}{2.4pt}
\begin{tabular}{@{}lrrr@{}}
\toprule
Method & PQ $\uparrow$ & Lost (\%) $\downarrow$ & Spurious / img $\downarrow$ \\
\midrule
Majority, class first (void first) & 93.54 & 24.0 & 0.76 \\
Majority, class first (void last) & \textbf{94.67} & 22.6 & 1.41 \\
Majority pooling & 93.51 & 24.2 & \textbf{0.59} \\
Majority pooling (void last) & \underline{94.66} & 22.2 & 1.47 \\
Majority pooling (ties to highest id) & 94.59 & 20.8 & 1.72 \\
One-hot bilinear + argmax & 93.40 & 23.7 & \underline{0.66} \\
nnU-Net per-label & 94.22 & 22.9 & 1.12 \\
nnU-Net per-label (void last) & 94.27 & 21.7 & 1.40 \\
Smallest-region priority & 92.15 & \underline{4.00} & 2.94 \\
Greedy rescue & 92.94 & 4.73 & 1.83 \\
$\kappa{=}4/8$, no repair & 93.56 & 20.1 & 0.81 \\
$\kappa{=}4/8$, repair & 93.56 & 20.1 & \textbf{0.59} \\
MatchDown, no repair & 93.01 & \textbf{1.29} & 2.52 \\
MatchDown (ours) & 92.96 & \textbf{1.29} & 1.70 \\
\bottomrule
\end{tabular}
\end{table}

\begin{table}[t]
\centering\footnotesize
\caption{\textbf{Reconstruction PQ on COCO-A} (1,000 images, all 133 classes) at factors 2, 4 and 8, and the share of all components, void and crowd included, lost at factor 4. Smallest-region priority gives each cell the smallest region covering at least a quarter of it. Keeping more regions lowers PQ at factors 4 and 8, because many of the kept regions are not reachable (\cref{prop:ceiling}). Best in bold, second best underlined.}
\label{tab:cocoa_pq}
\setlength{\tabcolsep}{3.5pt}
\begin{tabular}{@{}lrrrr@{}}
\toprule
 & \multicolumn{3}{c}{PQ $\uparrow$} & Lost (\%) $\downarrow$ \\
\cmidrule(lr){2-4}
Method & $f{=}2$ & 4 & 8 & $f{=}4$ \\
\midrule
Nearest (cell center) & 93.65 & \underline{87.92} & \underline{78.21} & 76.7 \\
Majority pooling & \textbf{93.81} & \textbf{88.58} & \textbf{78.86} & 86.7 \\
One-hot bilinear + argmax & \underline{93.69} & 87.08 & 75.36 & 75.8 \\
Smallest-region priority & 92.51 & 86.50 & 77.30 & 79.1 \\
Greedy rescue & 93.28 & 85.31 & 71.33 & \underline{19.3} \\
MatchDown, no repair & 93.35 & 85.23 & 70.86 & \textbf{17.1} \\
MatchDown (ours) & 93.32 & 84.90 & 70.05 & \textbf{17.1} \\
\bottomrule
\end{tabular}
\end{table}

\paragraph{Reconstruction quality.}
\Cref{tab:cocor2,tab:cocoa_pq} report the panoptic quality of the reconstructed targets.
At factor 2, MatchDown loses 336 evaluated regions on COCO-R at a PQ cost of 0.58 points against class-first majority pooling, and greedy rescue loses 1,228 at the same PQ cost.
At larger factors, keeping regions that are not reachable lowers recognition quality: on COCO at factor 8, $\kappa{=}1/8$ and $2/8$ lower PQ by 1.29 and 0.69 points against class-first majority, almost entirely through RQ (RQ terms $-1.13$ and $-0.64$, SQ terms $-0.16$ and $-0.05$).

\label{sec:supp:kappa}
\begin{table}[t]
\centering\footnotesize
\caption{\textbf{Coverage threshold $\kappa$.} Evaluated regions lost, as a fraction of those lost by class-first majority pooling on the same images. Uncovered-region bound: regions with no cell covered to $\kappa$ and no fixed majority cell, which every $\kappa$-restricted labeling loses (\cref{sec:supp:kappatheory}). Unweighted: the unweighted construction on the thresholded support graph. The tiered construction, with the regions already kept by uncovered cells ranked last, attains the exact minimum of \cref{thm:kappa} on every image, checked against the minimum computed separately; on Cityscapes at $\kappa{=}1/8$ that minimum is 0.5621.}
\label{tab:kappa}
\setlength{\tabcolsep}{5pt}
\begin{tabular}{@{}llrr@{}}
\toprule
Dataset & $\kappa$ & Uncovered-region bound & Unweighted \\
\midrule
Cityscapes ($f{=}8$) & $1/8$ & 0.5589 & 0.5627 \\
 & $2/8$ & 0.7254 & 0.7274 \\
 & $4/8$ & 0.9816 & 0.9816 \\
COCO ($f{=}8$) & $1/8$ & 0.8280 & 0.8303 \\
 & $2/8$ & 0.9149 & 0.9159 \\
 & $4/8$ & 0.9988 & 0.9988 \\
COCO-R ($f{=}2$) & $4/8$ & 0.8396 & 0.8396 \\
\bottomrule
\end{tabular}
\end{table}

\paragraph{Coverage threshold.}
The coverage threshold $\kappa$ trades the number of kept regions against their overlap with the cells they hold, and \cref{thm:kappa} gives the minimum loss at each $\kappa$ (\cref{tab:kappa}).
On 900 Cityscapes training images at factor 8, $\kappa{=}2/8$ raises class iIoU by 0.36 points over class-first majority pooling (95\% interval lower end 0.20), with 27\% fewer lost regions, comparable mIoU and 7.5\% more spurious adjacencies; $\kappa{=}1/8$ removes 44\% of majority's losses.
At factor 2 on COCO-R, $\kappa{=}4/8$ without repair loses 16\% fewer evaluated regions than class-first majority (5,231 against 6,230) with PQ 0.016 points higher (\cref{tab:cocor2}).
On the 1,000 COCO-A images at factor 4, $\kappa{=}2/8$ with repair loses 7.6\% fewer evaluated regions than one-hot bilinear interpolation with argmax, with 1.09 points higher reconstruction PQ (95\% interval $[+0.53,+1.65]$) and 26\% fewer spurious adjacencies; against majority pooling, it loses 12.9\% fewer regions at 0.41 points lower PQ.

\paragraph{Repair, binary topology and run time.}
\label{sec:supp:labelcmp}
An exact integer program (HiGHS) found the fewest spurious adjacent pairs at maximum survival on 200 random $64{\times}64$ BSDS crops: the repair attains it on 92.0 to 92.5\% of the crops, with a mean gap of 0.09 pairs and a maximum of 3, against 0.19 to 0.20 for the assignment alone.
The program omits the component guard, so the gap is conservative.
On 100 binary masks (50 BSDS, 50 COCO) at factors 2, 4 and 8, our reimplementation of Chen and Peng's integer program \cite{chen2024topology} preserves binary topology exactly whenever it returns a solution, but it returns none in 22 of the 300 cases: 21 are infeasible and one reaches the time limit, which does not establish infeasibility.
We test Hall's condition on the program's own support graph, which differs from ours: its components are 8-connected in the foreground and 4-connected in the background, the mask is padded with a background ring of $f$ pixels, and a component may take any cell whose block, dilated by $\lfloor f/4\rfloor$ pixels, holds one of its pixels.
Because the program gives every component a cell of its own, a Hall violation on this graph makes it infeasible.
Four of the infeasible cases violate the condition, and in all four the deficient components are background pieces enclosed by foreground; the other 17 infeasible cases and the time-out satisfy it, so the 17 fail for other constraints of the program, and no solved case violates the condition.
MatchDown always returns a labeling and preserves binary topology exactly on 60 to 88\% of the masks; it loses 0 to 11 of 170 BSDS components and 0 to 21 of 300 COCO components, against 27 to 58 and 74 to 110 for majority pooling, with a mean IoU at most 0.023 lower.
\Cref{tab:runtime} gives the run time.
\begin{table}[t]
\centering\footnotesize
\caption{\textbf{Run time} on one core of an Intel Xeon Platinum 8468: median over 120 inputs at factor 4 (five repeats) relative to majority pooling, and a full $1024{\times}2048$ Cityscapes frame at factor 8. Time grows about linearly with image size (log--log slopes 0.92 to 1.01). The target is built once before training; training and inference are unchanged.}
\label{tab:runtime}
\setlength{\tabcolsep}{4pt}
\begin{tabular}{@{}lrr@{}}
\toprule
Method & Relative time & Cityscapes frame \\
\midrule
Majority pooling & 1.0$\times$ & 0.69\,s \\
MatchDown & 1.7--1.9$\times$ & 1.29\,s \\
$\kappa$, no repair & 2.0$\times$ & 1.2\,s \\
$\kappa$, with repair & 2.8--3.1$\times$ & 2.0\,s \\
\bottomrule
\end{tabular}
\end{table}

\begin{table}[t]
\centering\footnotesize
\caption{\textbf{Where greedy rescue fails}, BBBC039 at factors 8 to 32 (197 images, 23,615 nuclei), label level. Bound (the lower bound of \cref{thm:certificate}), rescue and MatchDown-P: nuclei lost (\%); $\times$: rescue's loss as a multiple of the lower bound; $\ge$1 cell: nuclei of at least $f^2$ pixels lost by rescue (MatchDown-P loses none); gap: oracle count-error difference, rescue minus MatchDown-P, per image; images: share of images on which rescue loses more nuclei than MatchDown-P. MatchDown-P never loses more nuclei than rescue on any image.}
\label{tab:rescuefactor}
\setlength{\tabcolsep}{2.2pt}
\begin{tabular}{@{}rrrrrrrr@{}}
\toprule
$f$ & Bound & Rescue & $\times$ & MatchDown-P & $\ge$1 cell & Gap & Images (\%) \\
\midrule
8 & 0.025 & 0.182 & 7.2$\times$ & 0.025 & 0 & $+0.18$ & 13.2 \\
16 & 0.271 & 0.885 & 3.3$\times$ & 0.271 & 3 & $+0.71$ & 32.5 \\
24 & 0.648 & 3.231 & 5.0$\times$ & 0.648 & 55 & $+3.10$ & 86.3 \\
32 & 1.266 & 7.948 & 6.3$\times$ & 1.266 & 15 & $+8.01$ & 98.5 \\
\bottomrule
\end{tabular}
\end{table}

\paragraph{Where greedy rescue fails.}
On BBBC039 (\cref{tab:rescuefactor}), MatchDown-P loses exactly the lower bound at every factor from 8 to 32 and never more nuclei than greedy rescue on any image.
Rescue loses 3.3 times the lower bound at factor 16 and 6.3 times at factor 32, where its oracle count error is 9.5 nuclei per image against 1.5 for MatchDown-P and it loses more nuclei than MatchDown-P on 98.5\% of the images.

\section{Training results}
\label{sec:supp:trainres}
\begin{table*}[t]
\centering\footnotesize
\caption{\textbf{Affinity head on BBBC038 at factor 8} (670 images, 29,461 nuclei; 9 seeds $\times$ 3 folds, held-out predictions). Count error at the default thresholds; then differences from the majority model with 95\% confidence intervals, at cross-validated thresholds and at the equal-FP point (each model's thresholds chosen so that its false positives per image do not exceed the majority model's); detected: share (\%) of the 2,587 nuclei missing from the majority target that the model detects, cross-validated. $^\dagger$From the second set of runs, whose majority model has a count error of 5.95 (\cref{tab:controls}); its differences are taken against that model. Of the MatchDown model's count-error reduction, 89\% comes from nuclei that the majority model misses and the MatchDown model detects; the mean IoU bound of the 2,587 nuclei is 0.30 (\cref{sec:exp:ceiling}). Lost nuclei: \cref{tab:bbbc038star}. Best in bold, second best underlined, in the count-error and detected columns.}
\label{tab:training}
\setlength{\tabcolsep}{2.6pt}
\begin{tabular}{@{}lrllllr@{}}
\toprule
 & & \multicolumn{2}{c}{$\Delta$count err.~$\downarrow$} & $\Delta$\AccF~$\uparrow$ & $\Delta$centroid F1~$\uparrow$ & \\
\cmidrule(lr){3-4}
Target & Count err.~$\downarrow$ & cross-validated & equal FP & cross-validated & cross-validated & Detected (\%) $\uparrow$ \\
\midrule
Majority pooling & 5.93 & ref. & ref. & ref. & ref. & 31.4 \\
Cell center$^\dagger$ & 4.64 & $-1.01$ \scriptsize$[-1.34,-0.70]$ & $-1.00$ \scriptsize$[-1.34,-0.70]$ & $+0.022$ \scriptsize$[+0.015,+0.030]$ & $+0.016$ \scriptsize$[+0.011,+0.021]$ & -- \\
Greedy rescue & \underline{3.93} & $-1.43$ \scriptsize$[-2.46,-0.89]$ & $-0.51$ \scriptsize$[-1.47,+0.23]$ & $-0.010$ \scriptsize$[-0.016,-0.004]$ & $+0.023$ \scriptsize$[+0.018,+0.028]$ & 53.0 \\
Soft labels \cite{alcover2026soft} & 6.46 & $-1.64$ \scriptsize$[-2.65,-1.10]$ & $-1.38$ \scriptsize$[-2.40,-0.86]$ & $-0.082$ \scriptsize$[-0.106,-0.056]$ & $+0.019$ \scriptsize$[+0.013,+0.026]$ & \textbf{60.4} \\
Full-res.\ loss & \textbf{3.56} & $-1.34$ \scriptsize$[-2.40,-0.75]$ & $-0.57$ \scriptsize$[-1.56,+0.15]$ & $+0.018$ \scriptsize$[+0.011,+0.025]$ & $+0.028$ \scriptsize$[+0.024,+0.032]$ & \underline{59.4} \\
MatchDown (ours) & 4.03 & $-1.32$ \scriptsize$[-2.41,-0.75]$ & $-0.03$ \scriptsize$[-1.08,+0.81]$ & $-0.011$ \scriptsize$[-0.017,-0.004]$ & $+0.022$ \scriptsize$[+0.017,+0.027]$ & 52.8 \\
\bottomrule
\end{tabular}
\end{table*}

\begin{table*}[t]
\centering\footnotesize
\caption{\textbf{How the target keeps nuclei}, affinity head on BBBC038 at factor 8, default thresholds, from a second set of runs with the same protocol (so majority and MatchDown differ slightly from \cref{tab:training}). Masked: majority targets with the cells of the missing nuclei excluded from the loss. Dilated: every cell a nucleus touches is labeled with it. Differences from majority with 95\% confidence intervals. Best in bold (ties share the mark), second best underlined, in the lost, count-error and \AccF columns.}
\label{tab:controls}
\setlength{\tabcolsep}{6pt}
\begin{tabular}{@{}lrrllr@{}}
\toprule
Target & Lost (\%) & Count err. & $\Delta$count err. & $\Delta$\AccF & \AccF \\
\midrule
Majority pooling & 8.8 & 5.95 & ref. & ref. & \textbf{0.543} \\
Masked & 8.8 & 5.02 & $-0.93$ \scriptsize$[-1.17,-0.72]$ & $+0.000$ \scriptsize$[-0.004,+0.005]$ & \textbf{0.543} \\
Dilated & \textbf{0} & \underline{4.50} & $-1.45$ \scriptsize$[-1.89,-1.03]$ & $-0.234$ \scriptsize$[-0.249,-0.218]$ & 0.309 \\
MatchDown & \textbf{0} & \textbf{3.94} & $-2.00$ \scriptsize$[-2.40,-1.63]$ & $-0.012$ \scriptsize$[-0.017,-0.007]$ & \underline{0.531} \\
\bottomrule
\end{tabular}
\end{table*}

\begin{table*}[t]
\centering\footnotesize
\caption{\textbf{Star-convex head on BBBC038} (670 images, 29,461 nuclei; 9 seeds $\times$ 3 folds). Top: lost nuclei (\%), count error and \AccF at the default and cross-validated thresholds, and the share of the 2,587 nuclei missing from the majority target that the model matches at IoU${}>0.5$ (cross-validated). Bottom: differences in \AccF with 95\% confidence intervals. With default thresholds tuned for factor 8, the factor-2 strided model produces 27.7 false positives per image, and 5.7 with cross-validated thresholds. All models come from the second set of BBBC038 runs (\cref{sec:supp:training}). Best in bold, second best underlined, over all nine models.}
\label{tab:bbbc038star}
\setlength{\tabcolsep}{4pt}
\begin{tabular}{@{}llrrrrrr@{}}
\toprule
 & & & \multicolumn{2}{c}{Count err.~$\downarrow$} & \multicolumn{2}{c}{\AccF~$\uparrow$} & Missing nuclei \\
\cmidrule(lr){4-5}\cmidrule(lr){6-7}
Factor & Target & Lost (\%) $\downarrow$ & default & cross-val. & default & cross-val. & matched $\uparrow$ \\
\midrule
$f{=}8$ & Strided (StarDist) & 5.6 & 8.08 & 4.84 & 0.664 & 0.728 & 0.241 \\
 & Nearest (cell center) & 4.7 & \underline{5.04} & 5.78 & \underline{0.764} & 0.770 & 0.273 \\
 & Majority pooling & 8.8 & 6.34 & 6.50 & 0.746 & 0.750 & 0.152 \\
 & Greedy rescue & 0.35 & 5.14 & 4.77 & \textbf{0.765} & 0.772 & 0.331 \\
 & Dilated & \textbf{0} & 14.47 & 10.47 & 0.732 & 0.767 & 0.399 \\
 & MatchDown (ours) & \textbf{0} & \textbf{4.96} & \underline{4.10} & 0.761 & 0.770 & 0.347 \\
$f{=}4$ & Strided & 0.43 & 6.37 & 4.20 & 0.747 & 0.781 & 0.400 \\
 & MatchDown (ours) & \textbf{0} & 7.22 & \textbf{3.80} & 0.754 & \textbf{0.795} & \textbf{0.438} \\
$f{=}2$ & Strided (StarDist default) & \underline{0.04} & 22.90 & 4.30 & 0.713 & \underline{0.786} & \underline{0.421} \\
\midrule
\multicolumn{3}{@{}l}{Difference in \AccF} & \multicolumn{2}{c}{default} & \multicolumn{2}{c}{cross-validated} & \\
\cmidrule(lr){4-5}\cmidrule(lr){6-7}
\multicolumn{3}{@{}l}{Nearest $-$ strided, $f{=}8$} & \multicolumn{2}{c}{$+0.100$ \scriptsize$[+0.087,+0.113]$} & \multicolumn{2}{c}{$+0.042$ \scriptsize$[+0.034,+0.049]$} & \\
\multicolumn{3}{@{}l}{Majority $-$ strided, $f{=}8$} & \multicolumn{2}{c}{$+0.082$ \scriptsize$[+0.071,+0.093]$} & \multicolumn{2}{c}{$+0.022$ \scriptsize$[+0.015,+0.028]$} & \\
\multicolumn{3}{@{}l}{Dilated $-$ strided, $f{=}8$} & \multicolumn{2}{c}{$+0.068$ \scriptsize$[+0.054,+0.082]$} & \multicolumn{2}{c}{$+0.039$ \scriptsize$[+0.031,+0.046]$} & \\
\multicolumn{3}{@{}l}{MatchDown $-$ strided, $f{=}8$} & \multicolumn{2}{c}{$+0.097$ \scriptsize$[+0.084,+0.110]$} & \multicolumn{2}{c}{$+0.042$ \scriptsize$[+0.034,+0.049]$} & \\
\multicolumn{3}{@{}l}{MatchDown $-$ greedy rescue, $f{=}8$} & \multicolumn{2}{c}{$-0.005$ \scriptsize$[-0.010,+0.001]$} & \multicolumn{2}{c}{$-0.001$ \scriptsize$[-0.007,+0.004]$} & \\
\multicolumn{3}{@{}l}{Strided, $f{=}4$ $-$ strided, $f{=}8$} & \multicolumn{2}{c}{$+0.083$ \scriptsize$[+0.070,+0.097]$} & \multicolumn{2}{c}{$+0.053$ \scriptsize$[+0.046,+0.060]$} & \\
\multicolumn{3}{@{}l}{MatchDown $-$ strided, $f{=}4$} & \multicolumn{2}{c}{$+0.007$ \scriptsize$[+0.003,+0.011]$} & \multicolumn{2}{c}{$+0.014$ \scriptsize$[+0.010,+0.017]$} & \\
\multicolumn{3}{@{}l}{Strided, $f{=}2$ $-$ strided, $f{=}8$} & \multicolumn{2}{c}{$+0.049$ \scriptsize$[+0.033,+0.065]$} & \multicolumn{2}{c}{$+0.057$ \scriptsize$[+0.050,+0.065]$} & \\
\multicolumn{3}{@{}l}{Strided, $f{=}2$ $-$ MatchDown, $f{=}8$} & \multicolumn{2}{c}{$-0.048$ \scriptsize$[-0.055,-0.041]$} & \multicolumn{2}{c}{$+0.015$ \scriptsize$[+0.009,+0.021]$} & \\
\bottomrule
\end{tabular}
\end{table*}

\begin{table*}[!t]
\centering\footnotesize
\caption{\textbf{Sensitivity to the NMS search range}, star-convex head on BBBC038: held-out count error and \AccF at thresholds cross-validated with NMS IoU in $\{0.3,0.4,0.5\}$ (the range we report) or in $\{0.1,\dots,0.4\}$. The lower range raises the count error of several models, because on a few large images (up to $1040{\times}1388$ pixels) the selected thresholds produce hundreds of objects; the \AccF difference between MatchDown and strided at factor 8 stays positive, $+0.019$ $[+0.012,+0.026]$.}
\label{tab:nms}
\setlength{\tabcolsep}{4pt}
\begin{tabular}{@{}llrrrr@{}}
\toprule
 & & \multicolumn{2}{c}{Count err.~$\downarrow$} & \multicolumn{2}{c}{\AccF~$\uparrow$} \\
\cmidrule(lr){3-4}\cmidrule(lr){5-6}
$f$ & Target & $\ge0.3$ & $\ge0.1$ & $\ge0.3$ & $\ge0.1$ \\
\midrule
8 & Strided (StarDist) & 4.84 & 4.02 & 0.728 & 0.751 \\
 & Nearest (cell center) & 5.78 & 7.01 & 0.770 & 0.772 \\
 & Majority pooling & 6.50 & 6.45 & 0.750 & 0.755 \\
 & Greedy rescue & 4.77 & 4.27 & 0.772 & 0.775 \\
 & Dilated & 10.47 & 11.70 & 0.767 & 0.774 \\
 & MatchDown (ours) & 4.10 & 5.13 & 0.770 & 0.770 \\
4 & Strided & 4.20 & 3.61 & 0.781 & 0.788 \\
 & MatchDown (ours) & 3.80 & 4.49 & 0.795 & 0.792 \\
2 & Strided & 4.30 & 13.92 & 0.786 & 0.785 \\
\bottomrule
\end{tabular}
\end{table*}

\begin{figure*}[!t]
\centering
\includegraphics[width=\textwidth]{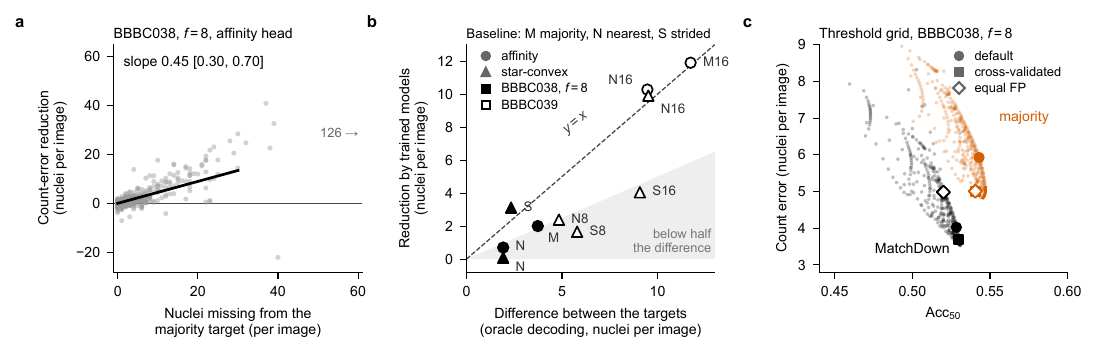}
\caption{\textbf{Where the count reduction of the affinity head comes from.} (a) BBBC038, $f{=}8$, default thresholds: per-image count-error reduction of the MatchDown model over the majority model (mean over nine seeds) against the nuclei missing from the majority target (one image, at 126, is off the axis). (b) Every comparison of MatchDown with a baseline on both datasets and heads: the count-error reduction of the trained models against the difference between the targets themselves under oracle decoding (each target used as the prediction); letters name the baseline, followed for BBBC039 by the factor. Points on the diagonal recover the whole difference; the affinity head does so at factor 16 and recovers about half of it at factor 8. (c) Count error against \AccF of the majority and MatchDown models at every point of the $19{\times}19$ threshold grid, with the default, cross-validated and equal-FP points marked (for majority the last two coincide). The MatchDown model has lower count error at every foreground threshold and lower \AccF at every threshold.}
\label{fig:mechanism}
\end{figure*}
\Cref{tab:training,tab:bbbc038star} give every BBBC038 model at the default and cross-validated thresholds, and \cref{tab:headline} summarizes them; \cref{fig:mechanism} shows where the count reduction of the affinity head comes from.
\Cref{tab:controls} separates keeping the nuclei from how their cells are assigned.
Excluding the cells of the missing nuclei from the loss recovers 0.93 nuclei per image of majority's count error at no \AccF cost, and MatchDown recovers a further 1.08 $[0.84, 1.32]$ at an \AccF cost of 1.3 points (paired difference between the two models).
A dilated target keeps every nucleus but enlarges each one by up to a ring of cells; its affinity model loses 23 \AccF points, and its star-convex model reaches 2.9 \AccF points below MatchDown at the default thresholds, with 12 more false positives per image.
At factor 4 the strided target loses only 0.43\% of the nuclei and recovers most of the \AccF difference to factor 8, at 3.3 more false positives per image.

\begin{figure}[t]
\centering
\includegraphics[width=\columnwidth]{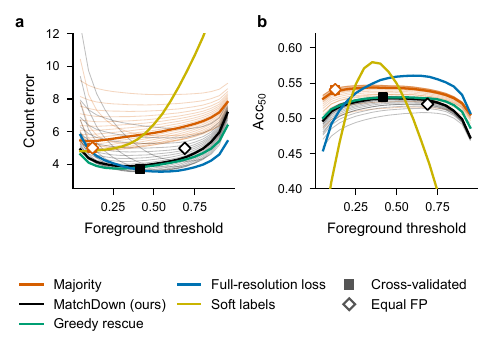}
\caption{\textbf{Operating curves of the affinity head} on BBBC038 at $f{=}8$ (means over 670 images and nine seeds): count error (a) and \AccF (b) against the foreground threshold, at the default affinity threshold (thick) and at the other 18 affinity thresholds (thin, majority and MatchDown only), for five of the targets of \cref{tab:training}. Squares mark the cross-validated thresholds (mean over the 27 selections) and diamonds the equal-FP points; for majority the two coincide. MatchDown has the lower count error at every foreground threshold and the lower \AccF at every threshold.}
\label{fig:operatingsupp}
\end{figure}
\paragraph{Threshold sensitivity.}
\Cref{fig:operatingsupp} shows the affinity models over the threshold grid.
For the star-convex head the selected NMS threshold lies at the lower edge of the grid, 0.3, for the strided and MatchDown models; \cref{tab:nms} extends the grid to 0.1.
Under the extension, the \AccF difference between MatchDown and strided shrinks from 4.2 to 1.9 points at factor 8 and from 1.4 to 0.4 at factor 4, and the count errors move in both directions because a few large images dominate them at the new points.
On BBBC038 the nearest model's cross-validated count error (5.78) is above its default (5.04), because thresholds selected on the tuning folds remove too many detections on the held-out fold.

\begin{table*}[t]
\centering\footnotesize
\caption{\textbf{Trained models on BBBC039} (197 images, 23,615 nuclei; 9 seeds $\times$ 3 folds, held-out predictions). Count error and \AccF at the default and cross-validated thresholds; \Acc, detected and matched at the default thresholds. Detected: share of the nuclei missing from the majority target (2,382 at factor 16, 1,363 at factor 8) that the model detects (\cref{sec:supp:metrics}); matched: share of the nuclei missing from the strided target (1,852 and 1,147) that the model matches at IoU${}>0.5$. Differences with intervals: \cref{tab:bbbc039diff}. Best in bold, second best underlined, within each head and factor.}
\label{tab:bbbc039}
\setlength{\tabcolsep}{6pt}
\begin{tabular}{@{}lrrrrrrr@{}}
\toprule
Target & \multicolumn{2}{c}{Count err.~$\downarrow$} & \Acc~$\uparrow$ & \multicolumn{2}{c}{\AccF~$\uparrow$} & Detected $\uparrow$ & Matched $\uparrow$ \\
\cmidrule(lr){2-3}\cmidrule(lr){5-6}
 & default & cross-val. & & default & cross-val. & & \\
\midrule
\multicolumn{8}{@{}l}{\textit{Affinity head, factor 16}} \\
Majority pooling & 15.86 & 13.05 & \underline{0.105} & \underline{0.472} & \underline{0.497} & 0.034 & 0.037 \\
Nearest (cell center) & \underline{14.23} & \underline{10.43} & \textbf{0.116} & \textbf{0.523} & \textbf{0.539} & \underline{0.133} & \underline{0.038} \\
MatchDown-P (ours) & \textbf{3.95} & \textbf{3.77} & 0.094 & 0.418 & 0.425 & \textbf{0.300} & \textbf{0.104} \\
\midrule
\multicolumn{8}{@{}l}{\textit{Star-convex head, factor 16}} \\
Strided (StarDist) & \underline{7.79} & \underline{5.62} & 0.594 & 0.816 & 0.799 & \underline{0.234} & 0.022 \\
Nearest (cell center) & 13.64 & 11.57 & \textbf{0.678} & \textbf{0.854} & \textbf{0.855} & 0.117 & \underline{0.081} \\
MatchDown-P (ours) & \textbf{3.74} & \textbf{3.35} & \underline{0.636} & \underline{0.831} & \underline{0.827} & \textbf{0.495} & \textbf{0.303} \\
\midrule
\multicolumn{8}{@{}l}{\textit{Star-convex head, factor 8}} \\
Strided (StarDist) & \underline{4.81} & \underline{4.29} & 0.669 & 0.844 & 0.847 & 0.128 & 0.019 \\
Nearest (cell center) & 5.54 & 5.51 & \textbf{0.704} & \textbf{0.883} & \underline{0.871} & \underline{0.216} & \underline{0.220} \\
MatchDown-P (ours) & \textbf{3.15} & \textbf{2.90} & \underline{0.683} & \underline{0.865} & \textbf{0.873} & \textbf{0.550} & \textbf{0.339} \\
\bottomrule
\end{tabular}
\end{table*}

\begin{table*}[t]
\centering\footnotesize
\caption{\textbf{Differences between targets on BBBC039} (197 images; 9 seeds $\times$ 3 folds, held-out predictions), at the default and the cross-validated thresholds, with 95\% confidence intervals from a bootstrap over seeds and images. Each row is MatchDown-P minus the named target; negative count-error differences favor MatchDown-P.}
\label{tab:bbbc039diff}
\setlength{\tabcolsep}{3pt}
\begin{tabular}{@{}llllll@{}}
\toprule
 & & \multicolumn{2}{c}{$\Delta$count error} & \multicolumn{2}{c}{$\Delta$\AccF} \\
\cmidrule(lr){3-4}\cmidrule(lr){5-6}
Head & Target & default & cross-val. & default & cross-val. \\
\midrule
Affinity, $f{=}16$ & minus majority & $-11.90$ \scriptsize$[-12.96,-10.89]$ & $-9.27$ \scriptsize$[-10.36,-8.24]$ & $-0.054$ \scriptsize$[-0.058,-0.050]$ & $-0.072$ \scriptsize$[-0.077,-0.067]$ \\
 & minus nearest & $-10.28$ \scriptsize$[-11.20,-9.40]$ & $-6.66$ \scriptsize$[-7.53,-5.80]$ & $-0.105$ \scriptsize$[-0.110,-0.100]$ & $-0.115$ \scriptsize$[-0.121,-0.108]$ \\
Star-convex, $f{=}16$ & minus strided & $-4.05$ \scriptsize$[-4.99,-3.22]$ & $-2.27$ \scriptsize$[-3.01,-1.58]$ & $+0.014$ \scriptsize$[+0.008,+0.022]$ & $+0.027$ \scriptsize$[+0.019,+0.037]$ \\
 & minus nearest & $-9.90$ \scriptsize$[-10.74,-9.11]$ & $-8.22$ \scriptsize$[-9.16,-7.32]$ & $-0.023$ \scriptsize$[-0.027,-0.018]$ & $-0.028$ \scriptsize$[-0.034,-0.023]$ \\
Star-convex, $f{=}8$ & minus strided & $-1.66$ \scriptsize$[-2.24,-1.11]$ & $-1.38$ \scriptsize$[-1.85,-0.92]$ & $+0.021$ \scriptsize$[+0.016,+0.025]$ & $+0.026$ \scriptsize$[+0.021,+0.031]$ \\
 & minus nearest & $-2.39$ \scriptsize$[-6.60,-0.70]$ & $-2.61$ \scriptsize$[-7.44,-0.12]$ & $-0.019$ \scriptsize$[-0.023,-0.013]$ & $+0.003$ \scriptsize$[-0.005,+0.010]$ \\
\bottomrule
\end{tabular}
\end{table*}

\begin{figure*}[!t]
\centering
\includegraphics[width=\textwidth]{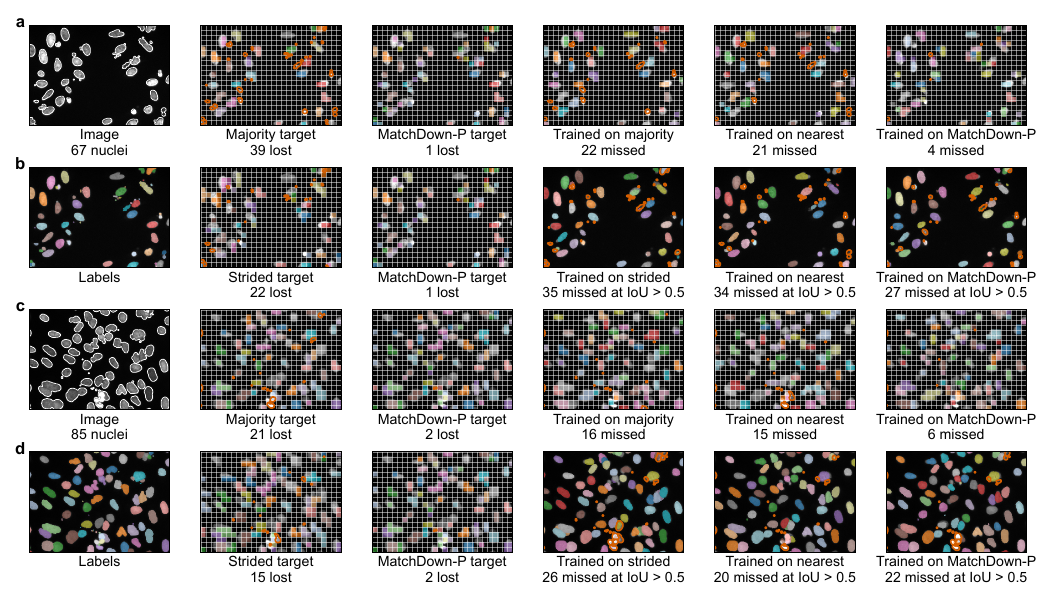}
\caption{\textbf{Two more held-out BBBC039 images at factor 16} ($320{\times}448$-pixel windows holding 67 and 85 of the images' 101 and 185 nuclei), as in \cref{fig:qual}, with the models trained on nearest targets added: (a, c) the affinity head, (b, d) the star-convex head. On the first image the affinity models trained on majority, nearest and MatchDown-P targets predict 52, 60 and 99 objects for 101 nuclei.}
\label{fig:qualsupp}
\end{figure*}
\begin{table*}[t]
\centering\footnotesize
\caption{\textbf{Greedy rescue and soft labels against MatchDown-P on BBBC039 at factor 16} (197 images; 9 seeds $\times$ 3 folds, held-out predictions, default thresholds; a separate set of runs with the BBBC039 protocol, in which the MatchDown-P models were trained again so that every difference is paired). Top: means; signed: mean predicted minus true count per image; FP50: false positives per image at IoU${}>0.5$; detected: share of the nuclei missing from the majority target (2,382) or from the rescue target (209) that the model detects; matched: share of the 1,852 nuclei missing from the strided target matched at IoU${}>0.5$. Bottom: differences, the named target minus MatchDown-P, with 95\% confidence intervals, and the difference between the targets themselves under oracle decoding. Best in bold within each head. One MatchDown-P star-convex run over-counts by 22.7 nuclei per image and widens that interval; because of it, the retrained MatchDown-P models' means differ from \cref{tab:bbbc039} (star-convex count error 4.76 here, 3.74 there; affinity 3.85 and 3.95).}
\label{tab:rescuesoft}
\setlength{\tabcolsep}{4pt}
\begin{tabular}{@{}llrrrrrrrr@{}}
\toprule
 & & & & & & & \multicolumn{2}{c}{Detected $\uparrow$} & Matched $\uparrow$ \\
\cmidrule(lr){8-9}\cmidrule(lr){10-10}
Head & Target & Count err.~$\downarrow$ & Signed & \Acc~$\uparrow$ & \AccF~$\uparrow$ & FP50 $\downarrow$ & majority-missing & rescue-missing & strided-missing \\
\midrule
Affinity & MatchDown-P (ours) & \textbf{3.85} & $-2.7$ & 0.094 & 0.418 & 48.1 & \textbf{0.300} & \textbf{0.308} & \textbf{0.104} \\
 & Greedy rescue & 4.45 & $-3.9$ & 0.101 & 0.449 & 43.4 & 0.289 & 0.262 & \textbf{0.104} \\
 & Soft labels \cite{alcover2026soft} & 15.36 & $-15.4$ & \textbf{0.105} & \textbf{0.472} & \textbf{33.0} & 0.051 & 0.065 & 0.042 \\
\midrule
Star-convex & MatchDown-P (ours) & 4.76 & $-1.7$ & 0.634 & 0.827 & 10.7 & \textbf{0.491} & \textbf{0.350} & 0.301 \\
 & Greedy rescue & \textbf{4.39} & $-3.6$ & \textbf{0.648} & \textbf{0.841} & \textbf{8.4} & 0.482 & 0.281 & \textbf{0.316} \\
\bottomrule
\end{tabular}

\medskip
{\footnotesize
\setlength{\tabcolsep}{1.8pt}
\begin{tabular}{@{}lllllll@{}}
\toprule
 & & & & \multicolumn{2}{c}{$\Delta$detected} & \\
\cmidrule(lr){5-6}
Target & $\Delta$count err. & $\Delta$\AccF & $\Delta$FP50 & majority & rescue & Oracle \\
\midrule
Affinity, rescue & $+0.60$ \scriptsize$[+0.26,+0.97]$ & $+0.031$ \scriptsize$[+0.027,+0.035]$ & $-4.7$ \scriptsize$[-5.3,-4.1]$ & $-0.01$ \scriptsize$[-0.02,+0.00]$ & $-0.05$ \scriptsize$[-0.08,-0.02]$ & $+0.71$ \\
Affinity, soft & $+11.51$ \scriptsize$[+10.62,+12.47]$ & $+0.055$ \scriptsize$[+0.050,+0.059]$ & $-15.1$ \scriptsize$[-16.2,-14.0]$ & $-0.25$ \scriptsize$[-0.26,-0.23]$ & $-0.24$ \scriptsize$[-0.30,-0.19]$ & -- \\
Star-convex, rescue & $-0.36$ \scriptsize$[-2.95,+1.09]$ & $+0.014$ \scriptsize$[+0.007,+0.022]$ & $-2.3$ \scriptsize$[-5.2,-0.8]$ & $-0.01$ \scriptsize$[-0.02,+0.01]$ & $-0.07$ \scriptsize$[-0.11,-0.02]$ & $+0.74$ \\
\bottomrule
\end{tabular}}
\end{table*}

\paragraph{BBBC039.}
\Cref{tab:bbbc039,tab:bbbc039diff} give the nine BBBC039 models and the differences between their targets, and \cref{fig:qualsupp} shows two more images with all trained models.
With cross-validated thresholds, the MatchDown-P affinity model has a count error 9.3 $[8.2, 10.4]$ nuclei per image lower than the majority model at an \AccF cost of 7.2 $[6.7, 7.7]$ points.
The majority and nearest affinity models select the corner of the grid (lowest foreground and highest affinity threshold), so no MatchDown-P threshold pair reaches their false-positive rate; the MatchDown-P models select interior points.
\Cref{tab:rescuesoft} adds greedy rescue and soft labels at factor 16.
The affinity model trained on rescue targets has a count error 0.60 $[0.26, 0.97]$ higher than the MatchDown-P model, 0.85 of the difference between the targets themselves, with 4.7 fewer false positives, \AccF 3.1 points higher, and 4.6 points fewer of the nuclei missing from its own target detected.
Soft labels raise the count error by 11.5 $[10.6, 12.5]$ and detect 24.8 points fewer of the nuclei missing from the majority target, at \AccF 5.5 points higher.
With the star-convex head, rescue and MatchDown-P have equal count error within the interval ($-0.36$ $[-2.95, +1.09]$), rescue's \AccF is 1.4 points higher, and its matched share of the nuclei missing from the strided target is within 0.015 of MatchDown-P's.

\end{document}